\documentclass[11pt, a4paper, logo, copyright]{googledeepmind}

\usepackage[numbers, sort&compress]{natbib} % Changed from authoryear to numbers per editor request

\usepackage{hyperref}
\usepackage{xurl}

\usepackage{cleveref}
\usepackage{url}
\usepackage{wrapfig,booktabs}
\usepackage{xcolor}         % colors
\usepackage{tikz}
\usetikzlibrary{patterns}
\usepackage{colortbl}
\usepackage{arydshln}
\usepackage{adjustbox}
\usepackage{multirow}
\usepackage{enumerate}
\usepackage{enumitem}
\usepackage{subfigure}
\usepackage{gensymb}
\usepackage{xspace} % for algorithm names
\usepackage{graphicx}
\usepackage{makecell}  % Include this in the preamble
\usepackage{minitoc}             % For creating the appendix TOC
\usepackage{makeidx}
\usepackage{booktabs}
\usepackage{bbm}
\usepackage{algorithm}
\usepackage{algorithmic}
\usepackage{colortbl}
\usepackage{psfrag}
\usepackage{adjustbox}
\usepackage{wrapfig}
\usepackage{xspace}
\usepackage{amssymb}
\usepackage{multirow}
\usepackage{afterpage}
\usepackage{float}
\usepackage[toc,page,header]{appendix}
\usepackage{minitoc}
\usepackage{algorithm}
\usepackage{algorithmic}
\usepackage{array}
\usepackage{amsmath}
\usepackage{pifont}
\usepackage{enumitem}

\definecolor{sigterm}{RGB}{0, 102, 204}  % Blue for significant
\definecolor{critterm}{RGB}{204, 0, 0} 

\newlist{myenum}{enumerate}{1}
\setlist[myenum]{leftmargin=2em,itemsep=0pt,topsep=0pt,partopsep=0pt}

\usepackage{placeins} % For FloatBarrier
\usepackage{titletoc} % Required for partial ToCs
\usepackage{array}
\usepackage{pifont} % for checkmarks and X marks

\definecolor{lightgray}{gray}{0.9}

\colorlet{darkgreen}{green!65!black}
\colorlet{darkblue}{blue!75!black}
\colorlet{darkred}{red!80!black}
\definecolor{lightblue}{HTML}{0071bc}
\definecolor{lightgreen}{HTML}{39b54a}
\definecolor{manyshot}{HTML}{6969ff}
\definecolor{medshot}{HTML}{f7c600}
\definecolor{fewshot}{HTML}{ff6969}
\definecolor{mypurple}{HTML}{412F8A}
\definecolor{myorange}{HTML}{fc8e62}

\definecolor{deemph}{gray}{0.55}

\definecolor{textgreen}{RGB}{57, 172, 57}
\definecolor{textred}{RGB}{200, 10, 10}
\definecolor{textgray}{RGB}{100, 100, 100}
\definecolor{visiongold}{RGB}{230, 184, 0}
\definecolor{speechpurple}{RGB}{204, 0, 255}
\definecolor{dataprep}{RGB}{38, 189, 128}
\definecolor{modeltraining}{RGB}{38, 189, 128}
\definecolor{backgroundcol}{RGB}{232, 230, 230}
\definecolor{gold}{rgb}{225, 215, 200} % Gold color
\definecolor{navyblue}{RGB}{40, 66, 200} % Navy blue color
\definecolor{orange}{RGB}{255,127,80} % Navy blue color
\definecolor{pink}{RGB}{219,112,147} % Navy blue color

\definecolor{baselinecolor}{gray}{.95}

\usepackage{pifont}% http://ctan.org/pkg/pifont

\usepackage{makecell, tabularx}
\newcolumntype{L}{>{\RaggedRight}X}
\usepackage{siunitx}

\title{Towards a Science of Scaling Agent Systems}

\correspondingauthor{ybkim95@mit.edu; dmduff@google.com; xliucs@google.com}

\reportnumber{}

\renewcommand{\today}

\author[1,3,$\dagger$]{Yubin Kim}
\author[1]{Ken Gu}
\author[3]{Chanwoo Park}
\author[2]{Chunjong Park}
\author[2]{Samuel Schmidgall}
\author[1]{A. Ali Heydari}
\author[1]{Yao Yan}
\author[1]{Zhihan Zhang}
\author[2]{Yuchen Zhuang}
\author[1]{Yun Liu}
\author[1]{Mark Malhotra}
\author[3]{Paul Pu Liang}
\author[3]{Hae Won Park}
\author[1]{Yuzhe Yang}
\author[1]{Xuhai Xu}
\author[1]{Yilun Du}
\author[1]{Shwetak Patel}
\author[1]{Tim Althoff}
\author[1,$\dagger$]{Daniel McDuff}
\author[1,$\dagger$]{Xin Liu}

\affil[1]{Google Research}
\affil[2]{Google DeepMind}
\affil[3]{Massachusetts Institute of Technology}
\affil[$\dagger$]{Corresponding Author}

\begin{abstract}
\textcolor{black}{\textit{Agents}, language model-based systems capable of reasoning, planning, and acting are widely adopted in real-world tasks, yet how their performance changes as these systems scale across key dimensions remains underexplored. We introduce quantitative \textit{scaling principles} for agent systems as a predictive model, capturing how performance varies with coordination, model capability, and measurable system and task factors. Across 260 configurations spanning six agentic benchmarks, five canonical architectures (Single-Agent and four Multi-Agent: Independent, Centralized, Decentralized, Hybrid), and three LLM families, we perform controlled evaluations standardizing tools, prompts, and compute to isolate architectural effects. The resulting model achieves a cross-validated $R^2{=}0.373$ across all six benchmarks ($R^2{=}0.413$ with a task-grounded capability metric). We identify a robust capability-saturation effect and additional patterns: \textbf{(1)} a coordination yields diminishing returns once single-agent baselines exceed certain performance; \textbf{(2)} tool-heavy tasks appear to incur multi-agent overhead; and \textbf{(3)} architectures without centralized verification tend to propagate errors more than those with centralized coordination. Relative performance change compared to single-agent baseline ranges from ${+}80.8\%$ on decomposable financial reasoning to ${-}70.0\%$ on sequential planning, demonstrating that architecture-task alignment determines collaborative success. The framework identifies the best-performing architecture for 87\% of held-out configurations and shows consistent relative architecture preferences on unseen frontier models. Agent effectiveness depends on alignment between coordination and task structure, and that mismatched coordination degrades the performance.}
\end{abstract}

\begin{document}

\maketitle

\newenvironment{Itemize}{
    \begin{itemize}[leftmargin=*]
    \setlength{\itemsep}{0pt}
    \setlength{\topsep}{0pt}
    \setlength{\partopsep}{0pt}
    \setlength{\parskip}{0pt}}
{\end{itemize}}
\setlength{\leftmargini}{9pt}

\begin{figure}[hbt]
    \centering
    \includegraphics[width=\textwidth]{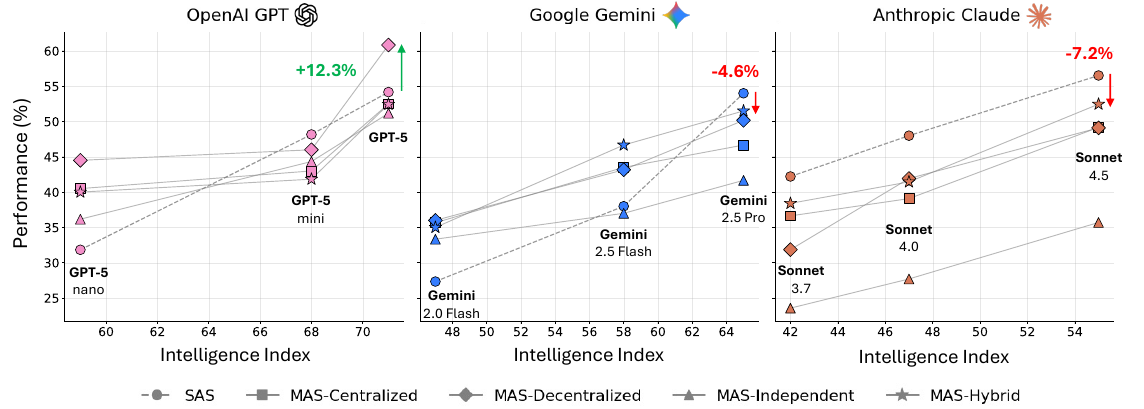}
    \caption{\textbf{Agent Scaling across model intelligence and system topologies.}
    \textcolor{black}{Average performance (\%) across the six agentic benchmarks} improves consistently with increasing model \emph{Intelligence Index} (see Appendix \ref{sec:appendix-intelligence-index}) across three LLM families (OpenAI, Google, and Anthropic) under five agent architectures. Single Agent System (\textbf{SAS}) serves as reference trajectories, while Multi Agent System (\textbf{MAS}) variants (Centralized, Decentralized, Independent, and Hybrid) reveal distinct scaling behaviors (see Table \ref{tab:agent_methods_complexity} for architecture comparisons).  
    All percentage deltas annotated in the figure indicate relative performance change of the best-performing MAS variant compared to the SAS baseline at the same Intelligence Index.}
    \label{fig:agentic_scaling}
\end{figure}

\section{Introduction}
\vspace{-0.1cm}

\textit{Agents} \citep{wang2024survey}, language model-driven systems that operate through iterative cycles of reasoning, planning, and acting, adapting their behavior based on environmental or tool-generated feedback, have achieved strong performance in diverse applications, from code generation \citep{zhang2024codeagent, yang2024swe}, web browsing \citep{wei2025browsecomp, yao2022webshop}, medical decision-making \citep{heydari2025anatomy, mcduff2025towards, kim2024mdagents}, finance \citep{yu2025finmem}, sustainability \citep{zhang2025towards}, to scientific discovery \citep{gottweis2025towards, mitchener2025kosmos}. As tasks become more complex and require long-horizon environmental interaction, multi-agent systems (MAS) have gained attention as a way to support task decomposition, parallel exploration, and verification. At the same time, concurrent works question whether multi-agent coordination outperforms single-agent systems (SAS), leaving the conditions under which MAS provides genuine benefits remain underexplored~\citep{tran2025multiagent,guo2024large, cemri2025multi,gao2025single,openhands2025single,cognition2025multi}. Despite rapid adoption, there remains no principled quantitative framework for predicting when adding agents improves performance and when it instead introduces coordination costs that degrade it. This gap leaves practitioners relying on heuristics, hindering both the emergence of a science of agent systems and, critically for real-world deployment, the ability to determine when multi-agent coordination provides genuine value over simpler single-agent alternatives.

To determine when multi-agent coordination provides benefit, we first establish which task categories require agentic capabilities. 
A necessary prerequisite is distinguishing between \textit{agentic} and \textit{non-agentic} evaluation paradigms. Expanding from the Agentic Benchmark Checklist (ABC) introduced in \citep{zhu2025establishing}, we characterize \textit{agentic tasks} as those requiring: (i) sustained multi-step interactions with an external environment, (ii) iterative information gathering under partial observability, and (iii) adaptive strategy refinement based on environmental feedback.

These characteristics differentiate tasks like web browsing~\citep{wei2025browsecomp}, financial trading~\citep{yu2025finmem}, software engineering~\citep{jimenez2023swe}, and interactive planning~\citep{dagan2024plancraft} from traditional static benchmarks, tasks solvable through single-shot reasoning without environmental feedback, which lack external environments, are fully observed, or require identical solution strategies~\citep{liu2023agentbench,kapoor2024agents}. This distinction matters because, while recent agentic benchmarks have emerged (e.g., SWE-Bench~\citep{jimenez2023swe}, $\tau^2$-Bench~\citep{barres2025t2bench}, Terminal-Bench~\citep{merrill2026terminal}), \textit{multi-agent system evaluations} have been conducted predominantly on non-agentic tasks, potentially providing misleading guidance about when collaboration provides value. This distinction is practically consequential: while LLMs achieve high accuracy on isolated code generation tasks like HumanEval~\citep{chen2021humaneval}, real-world deployment requires agentic capabilities such as iterative debugging, repository navigation, and adaptive strategy refinement as exemplified by interactive coding assistants (e.g., Cursor, Copilot Workspace). Multi-agent systems that show monotonic improvement with team size on static benchmarks (reaching 89\% on HumanEval with five agents) exhibit fundamentally different scaling behavior when evaluated on tasks requiring sustained environmental interaction, where coordination overhead and error propagation dynamics dominate. 

At its core, this distinction reflects a trade-off between context integration and diversity~\citep{du2023improving, hong2024metagpt}. Single-agent systems maximize context integration by maintaining a unified memory stream in which all reasoning steps share full access to prior history, enabling effectively constant-time access to global context. In contrast, multi-agent systems impose intrinsic information fragmentation~\citep{tran2025multiagent}: while parallel agents enable diverse exploration, they incur an unavoidable \textit{coordination tax} in which the global context must be compressed into inter-agent messages. This lossy communication increases synchronization overhead and cognitive load~\citep{malone1994interdisciplinary}, fundamentally altering the scaling behavior of collaboration.

The underlying dynamics explain this discrepancy: on agentic tasks, coordination overhead scales with interaction depth, agents operate on progressively divergent world states, and errors cascade through execution chains rather than being corrected through voting. Recent work has identified cases where single strong models match or exceed multi-agent systems~\citep{gao2025single}, yet the evaluation literature provides limited guidance on \textit{what factors} determine collaborative success, whether semantic diversity predicts team performance, how architectural choices shape coordination costs, or whether agents can detect and correct failures in extended interactions.

The problem is further compounded by rapid progress in frontier model capabilities. As base LLMs gain extended context windows, sophisticated tool use, and improved self-reflection, the unique value proposition of multi-agent collaboration becomes unclear. The answer likely depends on task characteristics and architectural choices that remain to be systematically quantified.

Two key challenges hinder progress toward principled multi-agent design. \textbf{First}, existing MAS evaluations compare architectures using different prompts, tools, or computational budgets, conflating architectural effects with implementation choices and precluding clean causal attribution. \textbf{Second}, evaluations focus exclusively on final accuracy metrics without examining process dynamics such as coordination overhead, error propagation, and information flow that determine whether collaboration succeeds or fails. We know from human team performance~\citep{mcgrath1964,lencioni2002} that team effectiveness depends on composition, coordination mechanisms, and member differentiation. Yet we lack comparable empirical understanding of how these principles translate to artificial agents, leaving practitioners without quantitative guidance for architecture selection.

To address these challenges, we present a controlled evaluation establishing the principles for agent coordination. Our experimental design isolates architectural effects by controlling for implementation confounds which maintains identical task prompts, tools, and computational budgets across all configurations, while systematically varying only coordination structure and model capability. \textcolor{black}{We evaluate five canonical architectures: Single Agent System (SAS) and four Multi-Agent variants (Independent, Centralized, Decentralized, Hybrid) instantiated across three major LLM families (OpenAI, Google, Anthropic) sampling models at varying capability tiers as quantified by an aggregate Intelligence Index (see Appendix \ref{sec:appendix-intelligence-index}), on six agentic benchmarks: (1) web browsing (BrowseComp-Plus \citep{chen2025browsecomp}), (2) financial analysis (Finance-Agent \citep{bigeard2025finance}), (3) game planning (PlanCraft \citep{dagan2024plancraft}), (4) realistic workplace tasks (Workbench \citep{styles2024workbench}), (5) software engineering (SWE-bench Verified \citep{jimenez2023swe}), and (6) terminal tasks (Terminal-Bench \citep{merrill2026terminal}). Across $N{=}260$ controlled configurations with matched compute, we derive a scaling principle across tested domains quantifying how performance emerges from empirically measured coordination properties.}

In contrast to prior claims that ``\textit{more agents is all you need}'', our evaluation reveals that the effectiveness of multi-agent systems is governed by quantifiable trade-offs between architectural properties and task characteristics. We establish a predictive framework using a mixed-effects regression model with empirical coordination metrics: efficiency (success/overhead ratio), error amplification factors, message density and redundancy as predictors, \textcolor{black}{achieving cross-validated $R^2{=}0.373$ across all six benchmarks ($R^2{=}0.413$ with a task-grounded capability metric), without dataset-specific parameters.} Critically, this framework generalizes beyond the fitted configurations in a restricted sense, where it predicts the best-performing architecture for 87\% of held-out task configurations, indicating relative architecture selection is more stable than absolute cross-domain performance prediction.

\textcolor{black}{Our analysis identifies three scaling patterns.} First, a \textit{tool-coordination trade-off} ($\beta{=}{-}0.096$, $p{=}0.002$): tool-heavy tasks (e.g., 16-tool business workflows) suffer from multi-agent coordination overhead, with efficiency penalties compounding as environmental complexity increases. Second, a \textit{capability ceiling} ($\beta{=}{-}0.236$, $p{=}0.004$): tasks where single-agent performance already exceeds 45\% accuracy experience negative returns from additional agents, as coordination costs exceed diminishing improvement potential. 
Third, we observe \textit{architecture-dependent error amplification}. Independent systems amplify trace-level errors $17.2\times$ through \textit{unchecked error propagation}, where individual mistakes cascade to the final output. Centralized coordination, however, contains this to $4.4\times$ by enforcing \textit{validation bottlenecks} that intercept errors before aggregation. Performance spans ${+}80.8\%$ relative improvement (structured financial reasoning under centralized coordination) to ${-}70.0\%$ degradation (sequential planning under independent coordination), demonstrating that architecture-task alignment, not number of agents, determines collaborative success. Optimal architectures vary systematically: decentralized coordination benefits tasks requiring parallel exploration of high-entropy search spaces (dynamic web navigation: ${+}9.2\%$), while all multi-agent variants universally degrade performance on tasks requiring sequential constraint satisfaction (planning: ${-}39\%$ to ${-}70\%$), where coordination overhead fragments reasoning capacity under fixed computational budgets. We translate these findings into quantitative architecture selection 
rules (Section~\ref{subsec:scaling}) achieving 87\% prediction accuracy on held-out configurations. The underlying mechanisms driving these patterns are interpretable: the tool-coordination trade-off arises because multi-agent systems fragment the per-agent token budget, leaving insufficient capacity for complex tool orchestration; the capability 
ceiling reflects that coordination overhead becomes a net cost when 
baseline performance is already high; and architecture-dependent error amplification stems from the presence or absence of validation bottlenecks that catch errors before propagation. These mechanistic insights enable 
practitioners to move from architectural heuristics to principled, measurement-driven deployment decisions.

Our primary contributions are:

\begin{itemize}

\item \textbf{Controlled evaluation of agent systems:} We establish a framework for comparing agent architectures, controlling for implementation confounds to isolate the effects of coordination structure. \textcolor{black}{Our framework spans 260 configurations across three LLM families and six diverse benchmarks, enabling controlled attribution of performance differences to architectural choices rather than stochastic variations.}

\item \textbf{Intelligence-Coordination alignment:} We characterize the non-linear relationship between foundational model capabilities and agentic performance. We demonstrate that while higher capability (Intelligence Index) yields consistent linear returns, these gains are not automatic; they strictly depend on architectural alignment. Without correct coordination structures, foundational improvements are often negated by coordination overhead.

\item \textbf{Quantitative scaling principles and architecture alignment:} We derive a regression model (\textcolor{black}{$R^2{=}0.373$ across all six benchmarks; $R^2{=}0.413$ with a task-grounded capability metric}) using empirical coordination metrics, efficiency ($E_c$), trace-level error amplification ($A_e^{\text{trace}}$), and redundancy ($\rho$) to quantify how performance emerges from the interplay of reasoning capability and task properties. This framework identifies fundamental limits on coordination, specifically a \textit{tool-coordination trade-off} ($\beta{=}{-}0.096$) where tool-heavy workflows suffer from coordination tax, and safety bounds where centralized verification reduces trace-level error amplification from $17.2\times$ to $4.4\times$. Using these mechanisms, we demonstrate that architecture selection is governed by measurable task features (e.g., decomposability) rather than simple agent scaling, achieving 87\% accuracy in predicting optimal architectures on held-out tasks.
\end{itemize}

\section{Related Work}

\paragraph{Multi-Agent Systems (MAS) versus Single-Agent Systems (SAS)}

Understanding the difference between single-agent and multi-agent systems remains central to characterizing architectural effects. Following \citet{tran2025multiagent} and \citet{guo2024large}, we define a \textbf{Single-Agent System} as one that features a solitary reasoning locus: all perception, planning, and action occur within a single sequential loop controlled by one LLM instance, even when employing tool use \citep{yao2023react}, self-reflection \citep{shinn2023reflexion}, or chain-of-thought (CoT) reasoning \citep{wei2022emergent}. Critically, self-reflection mechanisms do not constitute multi-agent collaboration, as they operate within a single decision-making locus \citep{weng2023llmagent}. A \textbf{Multi-Agent System} comprises multiple LLM-backed agents communicating through structured message passing, shared memory, or orchestrated protocols \citep{xi2025rise}. MAS architectures vary by topology: \textit{Independent} systems aggregate isolated outputs; \textit{Decentralized} enable peer-to-peer exchange \citep{du2023improving}; \textit{Centralized} route through orchestrators \citep{hong2024metagpt}; \textit{Hybrid} combine hierarchical control with lateral communication \citep{dang2025evolving}. MAS evaluation has moved beyond early assumptions of uniform superiority \citep{li2024more, qian2024scaling} towards a more differentiated understanding driven by domain complexity. Recent surveys characterize collaboration mechanisms across coordination protocols \citep{tran2025multiagent} and agent profiling patterns \citep{guo2024large}. However, there exist empirical challenges: \citet{gao2025single} show benefits diminish as base models improve, with frontier models often outperforming teams; \citet{cemri2025multi} identify 14 failure modes (Cohen's Kappa=0.88); \citet{zhang2025maas} achieve comparable performance at 6-45\% cost through dynamic architecture search; and \citet{anthropic2024multiagent} report agents consume 15$\times$ more tokens. Theoretical foundations from \citet{sumers2024cognitive} propose cognitive architectures contextualizing agents within AI's broader history. The question of \textit{when} multi-agent coordination provides value over single strong models with tool use remains empirically open, with \citet{qian2024scaling}'s proposed scaling laws showing no significant universal pattern \citep{wang2024survey}, motivating our systematic evaluation.

\paragraph{Agentic Tasks and Benchmarks}

We define \textit{agentic tasks} following \citet{zhu2025establishing} as requiring: (1) sustained multi-step environment interactions, (2) iterative information gathering under partial observability, and (3) adaptive strategy refinement from feedback, differentiating tasks like web browsing \citep{wei2025browsecomp, zhou2024webarena}, financial trading \citep{bigeard2025finance}, software engineering \citep{jimenez2023swebench}, and planning \citep{dagan2024plancraft} from static benchmarks. \textit{Non-agentic tasks} evaluate single-shot inference without environmental interaction: GSM8K \citep{cobbe2021gsm8k} (direct chain-of-thought math), MMLU \citep{hendrycks2020mmlu} (parametric knowledge), HumanEval \citep{chen2021humaneval} (specification-complete coding), and SQuAD \citep{rajpurkar2016squad} (single-pass comprehension). On non-agentic benchmarks, multi-agent systems show monotonic improvement through ensemble effects (89\% on HumanEval with five agents), as voting corrects errors without sequential compounding \citep{kapoor2024agents}. This distinction matters: in agentic settings, coordination overhead scales with interaction depth, agents operate on divergent world states (34\% overlap after 10 interactions), and errors cascade rather than cancel \citep{kapoor2024agents}. \citet{zhu2025establishing} introduce the Agentic Benchmark Checklist addressing flaws causing 100\% relative misestimation. Evolution spans \citet{liu2023agentbench}'s 8-environment evaluation (4k-13k responses) to specialized frameworks: \citet{jimenez2023swebench} (GitHub resolution), \citet{zhou2024webarena} (812 web tasks), \citet{xu2024theagentcompany} (30\% autonomous completion), and \citet{paglieri2024balrog} (vision-based RL). \citet{yao2023react} formalizes reasoning-acting synergy; \citet{weng2023llmagent} characterizes agents requiring planning, memory, and tools; \citet{kapoor2024agents} reveals narrow accuracy focus without cost metrics yields needlessly complex agents. \textcolor{black}{We note that established agentic benchmarks such as SWE-bench \cite{jimenez2023swe}, WebArena, and Tau-bench already embody these evaluation properties, as discussed in recent survey work \cite{yehudai2025survey}. Our contribution is not the formalization of these properties per se, but rather their systematic application as experimental controls across five coordination architectures and nine models from three LLM families, enabling the first quantitative characterization of how coordination benefit scales with model capability.} Tasks showing MAS advantages in single-shot settings often exhibit opposite patterns under genuine interaction, indicating architectural benefits are task-contingent, motivating our isolation of coordination effects across diverse agentic domains.

\paragraph{Scaling Laws and Coordination Mechanisms}

Understanding performance scaling in multi-agent systems requires distinguishing collaborative scaling from neural scaling laws. While neural scaling follows power laws requiring million-fold parameter increases for significant trends \citep{kaplan2020scaling}, collaborative scaling exhibits logistic growth patterns emerging at substantially smaller scales \citep{qian2024scaling}. \citet{chen2024compound} explore whether increased LLM calls alone drive performance, finding compound inference systems follow distinct scaling behaviors from single-model training. However, \citet{wang2024survey} note collaborative scaling shows no significant universal pattern, suggesting domain-specific rather than general laws. Coordination mechanisms critically determine whether collaboration amplifies or degrades performance: \citet{hong2024metagpt} introduce meta-programming workflows mitigating hallucination cascades; \citet{chen2023agentverse} demonstrate emergent behaviors through structured interactions; \citet{wu2024autogen} provide general multi-agent frameworks. Recent work reveals architecture-task alignment matters more than team size: \citet{zhang2025maas} achieve superior performance at 6-45\% cost through query-dependent configurations; \citet{dang2025evolving} show puppeteer orchestration improvements stem from compact cyclic structures; \citet{du2023improving} demonstrate peer-to-peer debate effectiveness depends on task decomposability, with \citet{smit2023should} further showing that multi-agent debate does not reliably outperform single-agent strategies such as self-consistency, suggesting benefits are highly task- and hyperparameter-sensitive. These findings collectively indicate coordination benefits arise from matching communication topology to task structure not from scaling the number of agents, establishing the foundation for principled architectural design rather than heuristic ``more agents is better'' approaches.

\section{Agent Systems and Tasks}
\label{sec:methods}

\subsection{System Definition}

Building on multi-agent system formalism \citep{zhu2025establishing, guo2024large}, an \textbf{agent system} $\mathcal{S} = (A, E, C, \Omega)$ consists of a set of agents $A = \{a_1, \ldots, a_n\}$ (where $n \geq 1$), a shared environment $E$, a communication topology $C$, and an orchestration policy $\Omega$. When $|A| = 1$, we refer to this as a Single-Agent System (SAS); when $|A| > 1$, a Multi-Agent System (MAS). Each agent $a_i$ perceives, reasons, and acts within the shared environment via iterative feedback.

Formally, each agent $a_i$ is defined as a tuple $S_i=(\Phi_i,\mathcal{A}_i,M_i,\pi_i)$, where:
\begin{itemize}
    \item $\Phi_i$ is the reasoning policy (typically an LLM)
    \item $\mathcal{A}_i = \{\text{ToolCall}(t, \theta) : t \in \mathcal{T}, \theta \in \Theta_t\}$ is the action space consisting of tool usage, where $\mathcal{T}$ is the set of available tools (e.g., web search, code execution) and $\Theta_t$ represents valid parameter configurations for tool $t$
    \item $M_i$ is the internal memory
    \item $\pi_i: \mathcal{H} \rightarrow \mathcal{A}_i$ is the decision function mapping observation histories to actions
\end{itemize}

The observation history space $\mathcal{H}$ contains sequences of action-observation pairs. The decision function $\pi_i$ is instantiated by the reasoning policy $\Phi_i$ (the LLM): given a history $h_{i,t}$, the LLM generates a reasoning trace and selects the next action. 

For instance, a history $h_{i,t} = [(\text{``search(query=`pandas')''}, \text{``Found 5 files''}), ...]$ is processed by $\Phi_i$ to produce the next tool call $\alpha_{i,t+1}$.

At timestep $t$, agent $a_i$ selects an action $\alpha_{i,t} \in \mathcal{A}_i$ according to:
\[
\alpha_{i,t}=\pi_i(h_{i,t}), \quad 
o_{i,t}=E(\alpha_{i,t}), \quad 
h_{i,t+1}=f_i(h_{i,t},\alpha_{i,t},o_{i,t}),
\]
where $E$ denotes the environment and $h_{i,0} = \{s_0\}$ contains the initial task specification. The history update function $f_i: \mathcal{H} \times \mathcal{A}_i \times \mathcal{O} \rightarrow \mathcal{H}$ appends the new action-observation pair to the agent's history: $h_{i,t+1} = f_i(h_{i,t}, \alpha_{i,t}, o_{i,t}) = h_{i,t} \oplus (\alpha_{i,t}, o_{i,t})$, subject to context window truncation when $|h_{i,t+1}| > \text{MAX\_TOKENS}$. This update mechanism applies uniformly to both SAS and MAS configurations. Communication between agents occurs through explicit message passing in the orchestration layer.

\vspace{0.4em}
\paragraph{Single-Agent System (SAS).}
A \emph{Single-Agent System} contains one reasoning locus ($|A|=1$ where $A$ is the agent set). All perception, reasoning, and action occur within a single sequential loop, producing computational complexity $O(k)$ where $k$ is the number of reasoning iterations. SAS has zero communication overhead and minimal memory $O(k)$, but limited capacity for decomposition or verification.

\vspace{0.4em}
\paragraph{Multi-Agent System (MAS).}

A \emph{Multi-Agent System} is an agent system $\mathcal{S}$ with $|A| > 1$, where agents interact through communication topology $C$ and orchestration policy $\Omega$.

Communication topology $C$ defines information flow patterns between agents:
\begin{itemize}
    \item \textbf{Independent}: $C = \{(a_i, a_{\text{agg}}) : \forall i\}$ (agent-to-aggregator only, no peer communication)
    \item \textbf{Centralized}: $C = \{(a_{\text{orch}}, a_i) : \forall i\}$ (orchestrator-to-agents only)  
    \item \textbf{Decentralized}: $C = \{(a_i, a_j) : \forall i,j, i \neq j\}$ (all-to-all topology)
    \item \textbf{Hybrid}: $C = C_{\text{centralized}} \cup C_{\text{peer}}$ (orchestrator plus limited peer-to-peer)
\end{itemize}

The orchestrator $\Omega$ (when present) determines: (i) how sub-agent outputs are aggregated (e.g., majority voting, weighted synthesis), (ii) whether the orchestrator can override sub-agent decisions, (iii) whether memory persists across coordination rounds, and (iv) termination conditions based on consensus or quality thresholds.

MAS architectures vary by how information and control propagate among agents, creating distinct trade-offs between computation, coordination, and parallelization. Table \ref{tab:agent_methods_complexity} formalizes these trade-offs using asymptotic notations over \textit{LLM calls}, \textit{sequential depth}, \textit{communication overhead}, and \textit{memory complexity}. We selected these five architectures to form a \textbf{structural ablation of coordination mechanisms}:

\begin{itemize}
    \item \textbf{Independent} isolates the effect of parallelism (ensemble) without communication.
    \item \textbf{Decentralized} introduces peer-to-peer information fusion without hierarchy.
    \item \textbf{Centralized} introduces hierarchical verification and bottleneck control.
    \item \textbf{Hybrid} examines the combination of hierarchy and lateral flexibility.
\end{itemize}

This design allows us to systematically attribute performance gains to specific coordination mechanics rather than generic ``multi-agent'' effects. Specific configurations include:
\begin{itemize}
    \item \textbf{Independent MAS:} $A = \{a_1, \ldots, a_n\}$, $\mathcal{C} = \{(a_i, a_{\text{agg}})\}$, $\Omega = \texttt{synthesis\_only}$. The \texttt{synthesis\_only} policy concatenates sub-agent outputs without cross-validation or majority voting; the aggregator performs no analytical comparison of responses, ensuring that any performance differences arise purely from parallel exploration rather than error correction. This achieves maximal parallelization but minimal coordination, suitable for ensemble-style reasoning.
    \item \textbf{Centralized MAS}: $A = \{a_{\text{orch}}, a_1,\ldots,a_n\}$, $C = \{(a_{\text{orch}}, a_i) : \forall i\}$, $\Omega = \text{hierarchical}$. A single orchestrator coordinates $r$ rounds across $n$ sub-agents ($O(rnk)$). Sequential depth equals $r$ while parallelization factor remains $n$. This design stabilizes reasoning but creates a bottleneck at the orchestrator.
    \item \textbf{Decentralized MAS}: $A = \{a_1,\ldots,a_n\}$, $C = \{(a_i, a_j) : \forall i,j, i \neq j\}$, $\Omega = \text{consensus}$. Agents communicate in $d$ sequential debate rounds ($O(dnk)$). Memory complexity is $O(dnk)$ as each agent stores its own debate history. This enables consensus formation through peer-to-peer discussion.
    \item \textbf{Hybrid MAS}: $A = \{a_{\text{orch}}, a_1,\ldots,a_n\}$, $C = \text{star} + \text{peer edges}$, $\Omega = \text{hierarchical} + \text{lateral}$. Combines orchestrated hierarchy with limited peer communication ($O((r+p) \cdot n \cdot k)$ where $p$ is the number of peer rounds). This inherits orchestrator control while enabling lateral exchange between agents.
\end{itemize}

\vspace{0.4em}
\paragraph{Communication vs. Coordination.}
We distinguish \emph{communication} (message passing between agents) from \emph{coordination} (strategic direction of agent activities). In centralized systems, coordination occurs through the orchestrator's task decomposition and progress monitoring, while communication involves passing findings between orchestrator and workers. In decentralized systems, communication and coordination are intertwined through debate rounds where agents both exchange information and collectively steer problem-solving direction.

Thus, SAS represents the minimal unit of agentic computation ($O(k)$), while MAS configurations explore the scaling frontier of coordination complexity, ranging from fully parallel and communication-free (Independent) to fully coupled with peer consensus (Decentralized). These configurations allow us to test whether performance gains arise from \emph{agent coordination and specialization} or merely from increased compute through ensembling. \textcolor{black}{Our taxonomy covers coordination patterns common in LLM-based agentic systems, focusing specifically on \emph{communication topology}, one of several orthogonal MAS design dimensions including agent specialization \citep{hong2024metagpt}, memory architecture, and aggregation strategy. Classical coordination mechanisms such as blackboard systems assume structured message formats rather than natural language, limiting their direct applicability to LLM-based agents \citep{guo2024large, xi2025rise}.}

\textcolor{black}{We formally define the task-level error rate as $E = 1 - P$, where $P$ is the fraction of tasks successfully resolved. The \emph{task-level} error amplification factor $A_e^{\text{task}} = E_{\text{MAS}} / E_{\text{SAS}}$ quantifies the relative error rate of a multi-agent system compared to its single-agent baseline; $A_e^{\text{task}} > 1$ indicates that coordination introduces net errors, while $A_e^{\text{task}} < 1$ indicates net error suppression. We additionally define a \emph{trace-level} error amplification factor $A_e^{\text{trace}}$ that measures how much extra computational work arises from inter-agent coordination failures, estimated from execution-trace token analysis (see Section~\ref{subsec:coord_eff}). Both metrics consistently show that architectures with verification mechanisms contain errors more effectively than independent coordination, though they differ in absolute magnitude ($A_e^{\text{task}} \approx 1.1$--$1.3$ vs.\ $A_e^{\text{trace}} \approx 4$--$17$) because they capture complementary aspects of error dynamics.}

\subsection{Agentic Tasks and Benchmarks}

Following and extending the framework of \citet{zhu2025establishing}, we operationalize a task $T$ as \textbf{agentic} when optimal performance \emph{substantially} benefits from adaptive interaction. Formally, if $\tau = \{(a_t, o_t)\}_{t=0}^T$ represents an interaction trajectory, then:
\[
\frac{\max_{\pi} \mathbb{E}[R(\tau)] - \max_{g} \mathbb{E}[R(g(x))]}{\max_{\pi} \mathbb{E}[R(\tau)]} > \delta,
\]

where $\pi$ represents an interactive policy, $g$ represents any single-forward-pass function, $R$ measures task success, $\delta$ is a task-dependent threshold, and the expectation is over task instances $x$ and stochastic environment dynamics. This definition captures tasks where interaction provides meaningful advantage over the best possible single-shot approach.

The expected return of an optimal policy thus hinges on sequential observation–action feedback, requiring agents to gather information, plan, and revise hypotheses under partial observability. Building on the Agentic Benchmark Checklist \citep{zhu2025establishing}, we formalize three necessary properties for agentic benchmarks:

\begin{itemize}
    \item \textbf{Sequential Interdependence}: Later actions depend on earlier observations; a one-shot policy cannot achieve high reward.  
    \item \textbf{Partial Observability}: Critical state information is hidden and must be acquired through active querying or tool use.  
    \item \textbf{Adaptive Strategy Formation}: The policy must update internal beliefs based on new evidence obtained through interaction.
\end{itemize}

\textcolor{black}{\noindent Benchmarks lacking these conditions (e.g., GSM8K, MMLU) evaluate static reasoning rather than agentic capabilities. (We note that ``agentic'' is defined relative to current model capabilities: GSM8K could be posed as agentic by providing calculator tools, though current LLMs do not \emph{require} such scaffolding; conversely, tasks that are agentic today, such as SWE-Bench, may become solvable via single-shot inference as models improve. Our evaluation focuses on tasks that \emph{currently} require multi-step interaction for non-trivial performance.)}

\vspace{0.4em}
\paragraph{Why Environment Feedback Matters.}
Real-world deployments such as coding assistants, financial analysts, and embodied robots operate under uncertainty and non-stationarity.  
Tasks solvable by direct prompting measure linguistic knowledge, whereas agentic benchmarks evaluate the process of intelligence: exploration, adaptation, and coordination.  
Hence, our benchmarks are chosen such that (i) base LLMs perform poorly in single-shot mode, and (ii) non-trivial performance requires multi-step environment interaction.

\vspace{0.4em}
\paragraph{Benchmark Design Principles.}
Extending the framework proposed by \citet{zhu2025establishing}, we introduce additional criteria to isolate \emph{architectural effects}:

\begin{itemize}
    \item \textbf{Controlled Tool Interface:} identical tool APIs and observation structures for all architectures to eliminate confounds from external feedback quality.
    \item \textbf{Controlled for Parametric Knowledge:} within each model family, evaluation emphasizes adaptive reasoning over memorized facts. Cross-family comparisons (Section~\ref{sec:results}) account for inherent knowledge base differences through baseline normalization.
    \item \textbf{Action–Observation Loop Length:} each benchmark enforces non-trivial trajectory length $L>3$ to ensure sequential reasoning.
    \item \textbf{Comparative Normalization:} scores are normalized to the best single-agent baseline, measuring coordination gain or loss.
\end{itemize}

\begin{table}[t!]
\centering
\textcolor{black}{\caption{Six agentic benchmarks used for evaluation.}}
\adjustbox{width=\textwidth}{%
\begin{tabular}{lll}
\toprule
\textbf{Benchmark} & \textbf{Task} & \textbf{Evaluation Design} \\
\midrule
BrowseComp-Plus (2025) & Web Browsing / Information Retrieval & Multi-website Information Location \\
\rowcolor{lightgray}
Finance-Agent (2025) & Finance & Entry-level Analyst Task Performance \\
Plancraft (2024) & Agent Planning & Minecraft Environment Planning \\
\rowcolor{lightgray}
WorkBench (2024) & Planning / Tool Selection  & Common business activities  \\
\textcolor{black}{SWE-bench Verified (2024)} & \textcolor{black}{Software Engineering} & \textcolor{black}{GitHub Issue Resolution} \\
\rowcolor{lightgray}
\textcolor{black}{Terminal-Bench (2025)} & \textcolor{black}{CLI Task Execution} & \textcolor{black}{System Admin / Security / ML Tasks} \\
\bottomrule
\end{tabular}%
}
\end{table}

\begin{table}[t!]
\centering

\caption{Architectural comparison of agent methods with objective complexity metrics. Computational complexity measured in terms of LLM calls, coordination overhead, and parallelization potential.}
\label{tab:agent_methods_complexity}

{\fontsize{7}{8.5}\selectfont
\begin{tabularx}{\textwidth}{p{2.4cm} >{\centering\arraybackslash}X >{\centering\arraybackslash}X >{\centering\arraybackslash}X >{\centering\arraybackslash}X >{\centering\arraybackslash}X}
\toprule
\textbf{Characteristic} & \textbf{SAS} & \makecell{\textbf{MAS} \\ (\textbf{Independent})} & \makecell{\textbf{MAS} \\ (\textbf{Decentralized})} &
\makecell{\textbf{MAS} \\ (\textbf{Centralized})}  & \makecell{\textbf{MAS} \\ (\textbf{Hybrid})} \\ \hline
\multirow{1}{*}[15pt]{\vspace{3em}\textbf{Interaction Type}}
& \includegraphics[width=0.65cm]{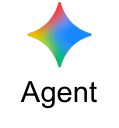}
& \includegraphics[width=2.3cm]{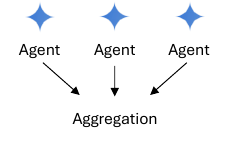}
& \includegraphics[width=2.4cm]{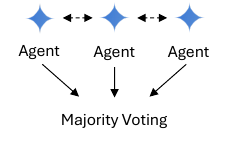}
& \includegraphics[width=2.65cm]{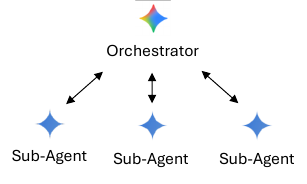}
& \includegraphics[width=2.65cm]{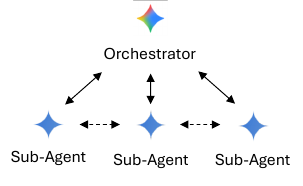} \\
\textbf{LLM Calls} & $O(k)$ & $O(nk) + O(1)$ & $O(dnk) + O(1)$ & $O(rnk) + O(r)$ & $O(rnk) + O(r) + O(p)$ \\
\textbf{Sequential Depth} & $k$ & $k$ & $d$ & $r$ & $r$ \\
\textbf{Comm. Overhead} & $0$ & $1$ & $d \cdot n$ & $r \cdot n$ & $r \cdot n + p \cdot m$ \\
\textbf{Parallelization Factor} & $1$ & $n$ & $n$ & $n$ & $n$ \\
\textbf{Memory Complexity} & $O(k)$ & $O(n \cdot k)$ & $O(d \cdot n \cdot k)$ & $O(r \cdot n \cdot k)$ & $O((r + p) \cdot n \cdot k)$ \\
\textbf{Coordination} & Sequential & Parallel + Synthesis & Sequential Debate & Hierarchical & Hierarchical + Peer \\
\textbf{Consensus} & - & Synthesis & Debate & Orchestrator & Orchestrator \\
\bottomrule
\end{tabularx}
}

\vspace{-0.3em}
\begin{flushleft}
{\fontsize{10}{11}\selectfont
\textbf{*} $k$ = max iterations per agent, $n$ = number of agents, $r$ = orchestrator rounds, $d$ = debate rounds, $p$ = peer communication rounds, $m$ = average peer requests per round. Communication overhead counts inter-agent message exchanges. Independent offers maximal parallelization with minimal coordination. Decentralized uses sequential debate rounds. Hybrid combines orchestrator control with directed peer communication.
}
\end{flushleft}

\end{table}

\section{Experiments \& Results}
\label{sec:results}

To establish quantitative scaling principles for agentic systems, we investigate three research questions:

\noindent\textbf{RQ1.} What factors determine agent system's performance (e.g., model capability, coordination architecture, task properties, their interactions)? \textcolor{black}{We systematically vary each factor across 260 configurations to quantify their individual and joint contributions.}

\noindent\textbf{RQ2.} Under what conditions does inter-agent coordination improve or degrade agent system's performance? We examine how task structure (e.g., decomposability, tool complexity, sequential dependencies) moderates the effectiveness of different architectures.

\noindent\textbf{RQ3.} Can we derive quantitative scaling principles that predict best agent architecture for a given task from measurable properties? We fit a regression model using empirical coordination metrics to test whether continuous properties outperform categorical architecture labels in explaining performance variance.

\begin{figure}[ht!]
  \centering
  \includegraphics[width=0.9\textwidth]{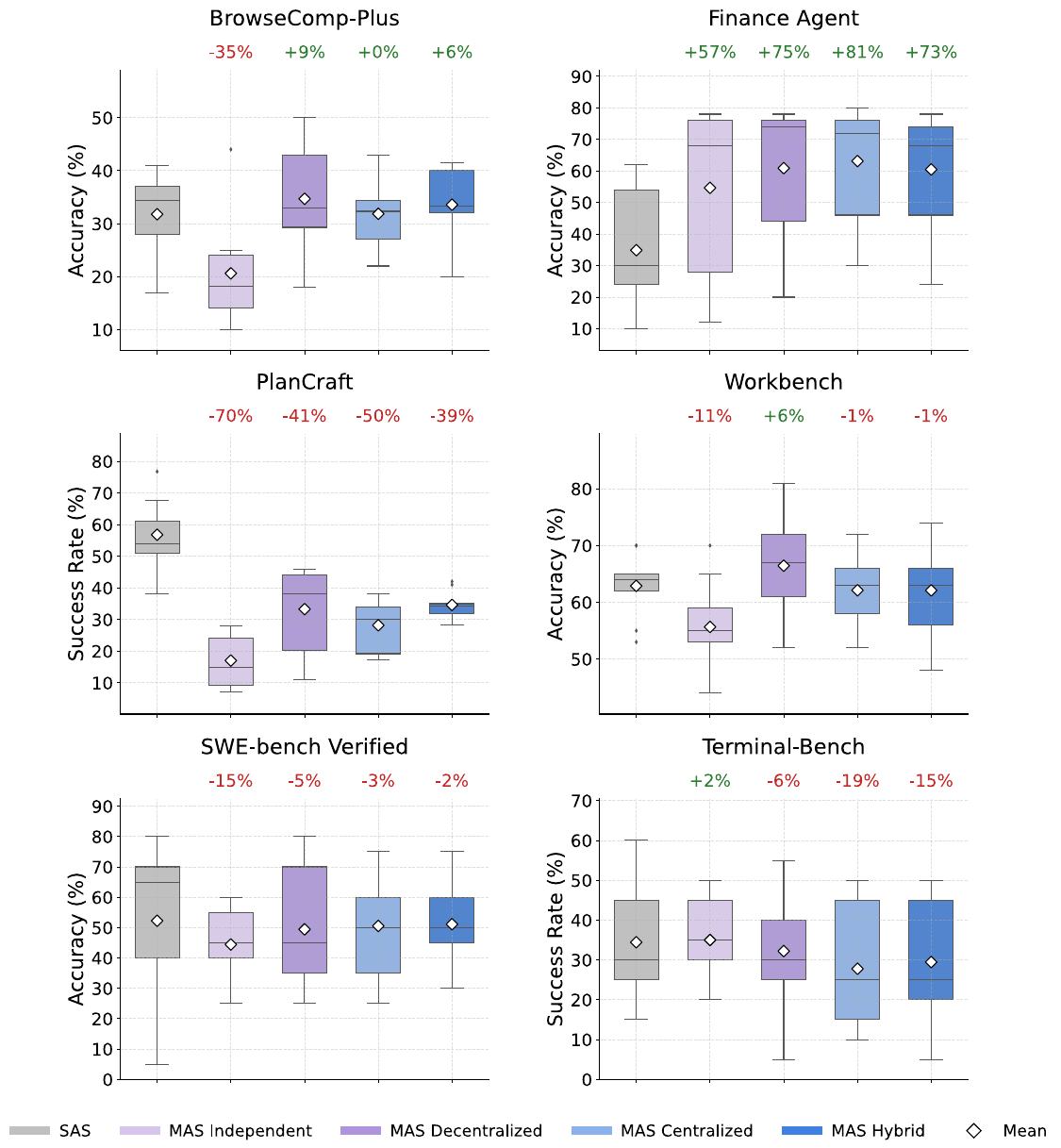}
  \caption{
  \textcolor{black}{\textbf{Comparative performance of agent systems across six agentic benchmarks reveals highly task-dependent scaling dynamics.}} \textcolor{black}{Box plots show distribution of performance (0--100\%).} Percentage annotations represent \textit{relative} improvement/degradation compared to SAS baseline: $(\text{mean}_{\text{MAS}} - \text{mean}_{\text{SAS}})/\text{mean}_{\text{SAS}} \times 100\%$. SAS serves as the reference baseline (shown without percentage annotation).
  \textbf{(a)} BrowseComp-Plus shows polarized results, with independent agents catastrophically underperforming relative to SAS (-35\%) while more structured coordination achieves modest gains. \textbf{(b)} Finance Agent demonstrates the strongest multi-agent benefits, with all MAS architectures substantially outperforming SAS (from +57\% to +80.8\%). \textbf{(c)} PlanCraft exhibits consistent degradation across all MAS variants (from -70\% to -39\%). \textbf{(d)} Workbench shows marginal effects (from -11 to +6\%). \textcolor{black}{\textbf{(e)} SWE-bench Verified shows slight degradation across all MAS architectures (from -15\% to -2\%), consistent with high single-agent baselines ($>$45\%) for most models. \textbf{(f)} Terminal-Bench shows mixed results: Independent achieves marginal gains (+2\%) while Centralized degrades (-19\%), reflecting the low tool count (2 tools) where coordination overhead is less justified.} 
  }
  \label{fig:boxplots}
  \end{figure}

\subsection{Setup}
  
\paragraph{Benchmarks.}
\textcolor{black}{We conducted 260 experiments across six benchmarks spanning deterministic to open-world task structures: \textbf{Workbench} (deterministic code execution and tool use with objective pass/fail criteria), \textbf{Finance Agent} (multi-step quantitative reasoning and risk assessment), \textbf{PlanCraft} (spatiotemporal planning under constraints), \textbf{BrowseComp-Plus} (dynamic web navigation, information extraction, and cross-page synthesis), \textbf{SWE-bench Verified} (real-world software engineering; GitHub issue resolution with 7 tools including bash, file editing, and test execution), and \textbf{Terminal-Bench} (diverse CLI tasks spanning system administration, security, and ML training; 2 tools). BrowseComp-Plus, Finance Agent, PlanCraft, and Workbench each contribute 45 configurations (9 models $\times$ 5 architectures); SWE-bench Verified and Terminal-Bench each contribute 40 configurations (8 models $\times$ 5 architectures, as Claude Sonnet 3.7 is deprecated). BrowseComp-Plus, Finance Agent, PlanCraft, and Workbench use 50--100 instances per configuration; SWE-bench Verified and Terminal-Bench use 20-instance subsets due to the computational cost of Docker-based evaluation (see Table~\ref{tab:s6_bootstrap} for bootstrap confidence intervals). The tool-count range spans $\{2,3,4,5,7,16\}$ across all six benchmarks.} BrowseComp-Plus exhibits the highest performance variability across experimental configurations (coefficient of variation $\sigma/\mu= 0.32$ computed across all 45 BrowseComp-Plus runs spanning architectures and model families, with Anthropic models contributing substantial variance due to lower absolute performance, where $\sigma$ is the standard deviation of success rates and $\mu$ is the mean success rate). By comparison, Workbench (CV=0.12), Finance Agent (CV=0.18), and PlanCraft (CV=0.21) show lower variability, indicating more stable performance across configurations.

\paragraph{LLMs and Intelligence Scaling.} We evaluate three LLM families across multiple model sizes, spanning externally standardized Intelligence Index values from 42 to 71 (a composite capability score integrating reasoning, coding, and knowledge benchmarks; see Appendix \ref{sec:appendix-intelligence-index}):
\begin{itemize}[leftmargin=1.5em, topsep=2pt, itemsep=1pt]
    \item \textbf{OpenAI:} \textit{GPT-5-nano, GPT-5-mini, GPT-5}
    \item \textbf{Google:} \textit{Gemini-2.0 Flash, Gemini-2.5 Flash, Gemini-2.5 Pro}
    \item \textbf{Anthropic:} \textit{Claude Sonnet 3.7, Claude Sonnet 4, Claude Sonnet 4.5}
\end{itemize}
\textcolor{black}{Claude Sonnet 3.7 was deprecated by Anthropic in February 2026 and is therefore unavailable for SWE-bench Verified and Terminal-Bench. On these two benchmarks, the Anthropic family includes Claude Sonnet 4 and Claude Sonnet 4.5, while the OpenAI and Google families remain unchanged, yielding 8 models per benchmark and a total of $N{=}260$ configurations.}
Strong consistency across families validates that coordination scaling follows model-agnostic principles: the maximum difference in architecture-specific scaling slopes between any two LLM families is $\Delta_{\max} = 0.023$ (computed as $\max_{i,j}|\hat{\beta}_{\text{arch},i} - \hat{\beta}_{\text{arch},j}|$ across families $i,j \in \{\text{OpenAI, Google, Anthropic}\}$), with coefficient of variation CV $< 0.02$ across families. To ensure computational fairness, we matched maximum total iterations between MAS and SAS systems: MAS configurations received equal computational budget through parallel agent processing (smaller per-agent iterations for $n$-agent teams), while SAS received proportionally more reasoning rounds to compensate for lack of parallel deliberation.
  
\paragraph{Agent Architectures and Complexity.}
We tested five coordination topologies: Single-Agent System (SAS) and four Multi-Agent System (MAS) variants: Independent, Centralized, Decentralized, and Hybrid. Rather than attempting exhaustive coverage of all possible architectures, we selected these four MAS configurations to form a structured ablation over two key coordination dimensions: 
(i) \textit{orchestrator presence} (hierarchical control vs. flat structure), and (ii) \textit{peer communication} (direct sub-agent interaction vs. isolated execution). Independent isolates pure ensemble effects without any inter-agent communication; Centralized introduces hierarchical verification through an orchestrator bottleneck; Decentralized enables 
peer-to-peer information fusion without hierarchy; and Hybrid combines both mechanisms (see Table \ref{tab:agent_methods_complexity} for formal complexity characterization). This design enables controlled
attribution of performance differences to specific coordination
mechanisms rather than generic ``multi-agent'' effects. Coordination complexity is parameterized by communication overhead: the total number of inter-agent message exchanges required per task, yielding empirical values ranging from 0\% (SAS) to 515\% (Hybrid), with Independent at 58\%, Decentralized at 263\%, and Centralized at 285\% relative to the single-agent baseline (see Table~\ref{tab:coord_metrics}).
  
\paragraph{Metrics and Validation.}
Primary outcome is task success/accuracy (domain-dependent: factual correctness for Finance Agent, task completion for Workbench, goal satisfaction for PlanCraft, page synthesis accuracy for BrowseComp-Plus). Secondary metrics include: (i) factual error rate $E$ via domain-specific validators (Cohen's $\kappa$ \citep{cohen1960coefficient}: Finance Agent $= 0.91$, Workbench $= 0.89$, PlanCraft $= 0.87$, BrowseComp-Plus $= 0.88$; exceeding 0.80, indicating strong inter-rater reliability); (ii) information gain $\Delta \mathcal{I}$ from pre- vs.\ post-coordination uncertainty proxies (see Eq.~\ref{eq:info_gain}); (iii) token-overlap structure across agent rationales, labeling tokens as unique (appearing in exactly one agent), shared (two or more agents), or contradictory (semantic opposition detected when BERTScore similarity $< 0.3$ between assertion pairs, i.e., $1 - \text{BERTScore} > 0.7$, following the dissimilarity threshold established by \citet{zhang2019bertscore}); (iv) efficiency metrics including success per 1,000 tokens and cost-normalized performance. All metrics are normalized per reasoning turn and per token to enable cross-architecture comparison. We select coordination metrics based on two criteria: (i) direct measurability from experimental traces without requiring ground-truth labels beyond task success, and (ii) coverage of distinct aspects of coordination--performance relationships identified in prior work \citep{cemri2025multi}. We excluded metrics requiring subjective human annotation (e.g., solution creativity) or those exhibiting high collinearity with included measures (e.g., total message count correlates $r > 0.92$ with overhead). Variance inflation factor (VIF) analysis confirmed no severe multicollinearity among retained predictors (all VIF $< 5$). Specifically:

\begin{itemize}[leftmargin=1.5em, topsep=2pt, itemsep=1pt]

\item \textbf{Coordination overhead} $O = (T_{\text{MAS}} - T_{\text{SAS}})/T_{\text{SAS}} \times 100\%$: captures computational cost, identified as a primary bottleneck in production multi-agent deployments.

\item \textbf{Message density} $c$ (inter-agent messages per reasoning turn): quantifies communication intensity, a key factor in coordination scaling.

\item \textbf{Redundancy rate} $R$ (mean cosine similarity of agent output embeddings): measures agent agreement, relevant for ensemble-based error correction.

\item \textbf{Coordination efficiency} $E_c = S/(T/T_{\text{SAS}})$ (success normalized by relative turn count): normalizes success by cost for deployment decisions.

\item \textbf{Error amplification} $A_e^{\text{trace}}$ (trace-level error propagation factor, estimated from execution-trace token analysis): quantifies how coordination failures compound through agent interactions. The complementary task-level metric $A_e^{\text{task}} = E_{\text{MAS}}/E_{\text{SAS}}$ is defined in Section~\ref{sec:methods}.
\end{itemize}
  
\subsection{Main Results}
\label{subsec:main_results}
  
\paragraph{MAS exhibits domain-dependence with architectural variation.}
Multi-agent systems show highly variable performance across task domains, depending on problem structure and architectural choices. On Finance Agent, MAS achieve substantial improvements: Centralized reaches \textbf{+80.8\%} (mean 0.631 vs.\ SAS 0.349), Decentralized achieves \textbf{+74.5\%} (0.609), and Hybrid reaches \textbf{+73.1\%} (0.604), driven by opportunities for distributed financial reasoning across multiple agents. On Workbench, multi-agent systems show minimal gains: Decentralized achieves \textbf{+5.6\%} (0.664 vs.\ SAS 0.629), while Centralized and Hybrid both slightly underperform at \textbf{-1.2\%}. On BrowseComp-Plus, improvements remain modest: Decentralized achieves \textbf{+9.2\%} (0.347 vs.\ SAS 0.318), with Centralized essentially flat at \textbf{+0.2\%}. Critically, PlanCraft exhibits universal performance degradation 
across all multi-agent architectures. Centralized declines to 
$-50.3$\% (0.282 vs.\ SAS 0.568), Decentralized to $-41.5$\% (0.332), Hybrid to $-39.1$\% (0.346), and Independent to $-70.0$\% (0.170). To understand this
contrast between Finance Agent's gains and PlanCraft's degradation, we examined execution traces from both domains. In PlanCraft, efficient single-agent trajectories follow direct execution paths. For example, crafting a \texttt{diorite\_wall}:
\begin{quote}
\texttt{Turn 1: search("diorite\_wall") $\rightarrow$ Recipe: 6 diorite in 2x3}\\
\texttt{Turn 2: move(diorite $\rightarrow$ crafting\_grid)}\\
\texttt{Turn 3: craft $\rightarrow$ Task complete}
\end{quote}
In contrast, centralized multi-agent systems decompose inherently sequential tasks into artificial subtasks:
\begin{quote}
\texttt{Agent 1: Research recipe} (redundant, since lookup is instantaneous)\\
\texttt{Agent 2: Check inventory} (redundant, since state is visible to all)\\
\texttt{Agent 3: Execute crafting} (the only necessary step)
\end{quote}
This unnecessary decomposition generates substantial coordination messages on average for tasks requiring only a few execution steps, consuming token budget on coordination rather than reasoning. Conversely, Finance Agent trajectories demonstrate when coordination provides genuine value. Single-agent execution exhibits sequential bottlenecks:
\begin{quote}
\texttt{Turn 1: web\_search("merger news") $\rightarrow$ Surface results}\\
\texttt{Turn 2: edgar\_search("filings") $\rightarrow$ Limited depth}\\
\texttt{Turn 3--7: Sequential exploration with insufficient breadth}
\end{quote}
Centralized coordination enables parallel information synthesis:
\begin{quote}
\texttt{Agent 1: Regulatory/news analysis}\\
\texttt{Agent 2: SEC filing research}\\
\texttt{Agent 3: Operational impact assessment}\\
\texttt{Orchestrator: Synthesize multi-source findings}
\end{quote}
The task's natural decomposability such as revenue, cost, and market factors can be analyzed independently which aligns with the coordination structure, yielding $+80.8$\% improvement. These trajectory patterns reveal the mechanistic basis for domain-dependence: coordination overhead becomes counterproductive when coordination complexity exceeds task complexity (PlanCraft), but provides substantial gains when tasks naturally decompose into parallel information streams (Finance Agent).

\textcolor{black}{On SWE-bench Verified, all MAS architectures show slight degradation relative to SAS (mean 0.522): Hybrid $-2.1\%$ (0.511), Centralized $-3.1\%$ (0.506), Decentralized $-5.4\%$ (0.494), and Independent $-14.9\%$ (0.444). This is consistent with the capability-saturation threshold: most models achieve single-agent baselines above 45\%, leaving limited room for coordination gains. On Terminal-Bench (SAS mean 0.344, below the threshold), results are mixed: Independent shows marginal gains ($+1.7\%$, 0.350) while Centralized degrades substantially ($-19.2\%$, 0.278), suggesting that the low tool count (2 tools) limits the benefit of orchestration-heavy architectures.}

\textcolor{black}{Aggregating across all six benchmarks and architectures, the overall mean MAS improvement is $-0.3$\% (95\% CI: [$-58.7$\%, $+77.2$\%]), reflecting substantial performance heterogeneity with high variance ($\sigma = 37.5$\%).} The performance range across MAS variants spans from $-70.0$\% (PlanCraft Independent) to $+80.8$\% (Finance Centralized), indicating that MAS do not provide universal benefits but rather domain-specific trade-offs.
  
\paragraph{Domain Complexity Moderates Coordination Efficacy.} \textcolor{black}{Empirical patterns across benchmarks reveal that domain complexity (refer to Appendix \ref{appendix:domain_complexity} for details) moderates MAS advantage: structured, decomposable domains show large gains while high-complexity sequential domains show consistent degradation.} The mechanism operates through fixed computational budgets (matched total tokens across MAS and SAS): in structured, decomposable domains (Finance Agent, moderate Workbench instances), agents complete local reasoning with residual capacity available for inter-agent communication. Here, inter-agent messages reduce variance through redundancy elimination and enable synthesis of partial solutions, producing large performance deltas (Finance: $+80.8\%$). Conversely, in high-complexity sequential domains (PlanCraft), intra-agent reasoning for constraint verification and state tracking consumes most available tokens before communication can occur; subsequent inter-agent messages then compress reasoning quality and produce strong negative returns (PlanCraft: $-39.0\%$ to $-70.0\%$).
  
\textcolor{black}{We characterize each benchmark by a domain complexity score $D \in [0,1]$ (Appendix~\ref{appendix:domain_complexity}), capturing the degree of sequential interdependence and empirical difficulty:} Workbench (0.000, minimal sequential constraints) shows positive MAS returns or minimal overhead, \textcolor{black}{SWE-bench Verified (0.255, decomposable engineering tasks) has low domain complexity but high single-agent baselines that trigger capability saturation,} Finance Agent (0.407, moderate decomposability) \textcolor{black}{and Terminal-Bench (0.414, diverse CLI tasks)} sit near the critical threshold, while PlanCraft (0.419, high sequential dependencies) and BrowseComp-Plus (0.839, dynamic state evolution) show degradation or minimal gains. Domain complexity alone does not fully predict MAS effectiveness. While low-complexity domains (Workbench, D = 0.00) show modest gains and high-complexity domains (BrowseComp-Plus, D = 0.84) show limited benefits, the critical factor is task decomposability: Finance Agent (D = 0.41) achieves +80.8\% gains through parallelizable subtask structure, whereas PlanCraft (D = 0.42) degrades by -70\% due to strict sequential dependencies despite similar complexity scores. This suggests that sequential interdependence, rather than complexity alone, determines coordination viability. Information gain $\Delta\mathcal{I}$ correlates with this pattern: Finance Agent (structured domain) exhibits strong information-value convergence ($r = 0.71$, $p < 0.001$), while PlanCraft (sequential constraints) shows weak correlation ($r = 0.18$, $p = 0.22$), indicating that agents in high-complexity domains exchange limited actionable information due to inherent sequential dependencies and state-space ambiguity.
  
\paragraph{Architecture-LLM Family Interactions Reveal Vendor-Specific Coordination Mechanisms.}
While domain complexity broadly moderates MAS effectiveness, the architecture-domain interaction reveals \emph{non-uniform} preferences even within similar complexity regimes: no single architecture dominates across all domains and vendors. Architecture effectiveness depends critically on domain structure: \texttt{Finance Agent} benefits most from Centralized (+80.8\%) and Decentralized (+74.5\%), \texttt{Workbench} from MAS-Decentralized (+5.6\%), and \texttt{BrowseComp-Plus} from MAS-Decentralized (+9.2\%). In degrading domains, architecture selection becomes a least-worst optimization: \texttt{PlanCraft} shows Hybrid as relatively best (-39.1\%) compared to MAS-Centralized (-50.3\%) and MAS-Independent (-70.0\%).
  
Family-specific coordination preferences emerge within improvement-positive domains. On \texttt{Finance Agent}, Anthropic's MAS-Centralized achieves +127.5\% (0.636 vs.\ 0.280 SAS), indicating conservative but stable coordination, whereas Google's MAS-Centralized reaches +164.3\% (0.740 vs.\ 0.280 SAS, averaging Centralized performance), suggesting stronger attention-mechanism alignment with hierarchical message exchange; OpenAI's MAS-Centralized achieves +69.9\% (0.79 vs.\ 0.465 SAS). On \texttt{Workbench}, where multi-agent overhead is less tolerable (efficiency degrades from $E_c = 0.466$ for SAS to $E_c = 0.074$ for Hybrid, the largest relative drop across benchmarks), Anthropic's best variant (MAS-Decentralized, +10.8\%) remains superior to Google (+9.5\%) and OpenAI (+8.6\%), reflecting relative efficiency in managing coordination costs. On \texttt{PlanCraft}, where all variants degrade, vendor preferences flatten: Anthropic shows maximum -54.5\% (MAS-Hybrid 0.31 vs.\ SAS 0.68), Google shows -25.3\% (best), and OpenAI shows -32.3\%, indicating that communication mechanisms cannot overcome fundamental sequential reasoning constraints.  While the precise mechanisms remain to be characterized, potential factors include differences in instruction-following fidelity, context utilization patterns, and inter-turn consistency that affect how agents interpret and respond to coordination messages. No vendor achieves universal multi-agent dominance; instead, each exhibits relative advantages in structured domains (Finance) that evaporate in sequential constraint-satisfaction domains (\texttt{PlanCraft}), indicating that multi-agent benefits are genuinely contingent on problem structure rather than generalizable across task types.

\begin{figure}[t!]
    \centering
    \includegraphics[width=\textwidth]{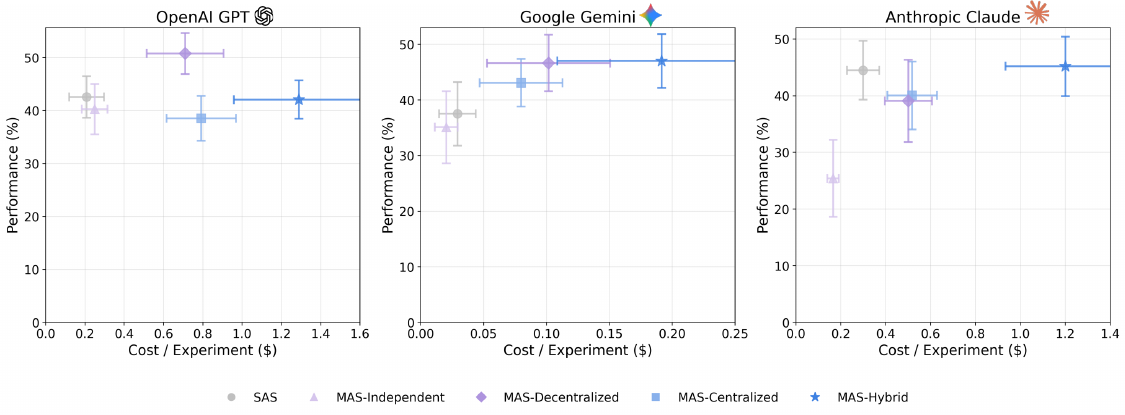}
    \caption{
      \textcolor{black}{\textbf{Cost–Performance Trade-offs Across Model Families and Architectures.} Data from BrowseComp-Plus, Finance Agent, PlanCraft, and Workbench (180 configurations with full cost tracking; SWE-bench Verified and Terminal-Bench use Docker-based evaluation with different cost structures).}
      Comparative analysis of single-agent and multi-agent architectures: Independent, Decentralized, Centralized, and Hybrid across three LLM families.
      Each point represents the mean agentic performance (\%) versus normalized cost per experiment (USD), with horizontal and vertical error bars denoting Standard Error of Mean (SEM) in cost and performance, respectively.
      The optimal coordination pattern differs across model families: OpenAI models show consistent gains from Centralized and Hybrid MAS configurations despite higher costs, suggesting stronger communication alignment;
      Google models display marginal MAS improvements but a clear efficiency plateau, indicating diminishing returns under lightweight coordination;
      and Anthropic models reveal higher variance and occasional MAS underperformance, reflecting sensitivity to coordination overhead.
      These cross-family discrepancies imply that \emph{the efficacy of multi-agent coordination is contingent on each model family’s intrinsic communication bandwidth and reasoning alignment}.
      Collectively, the results establish a family-dependent scaling principle linking coordination structure, economic efficiency, and emergent performance.}
    \label{fig:cost_performance}
\end{figure}

\subsection{Scaling principles}
\label{subsec:scaling}

The main results reveal substantial heterogeneity where agentic system performance ranges from $+80.8\%$ improvement to $-70\%$ degradation depending on task structure and coordination architecture. This variance correlates with measurable properties such as task decomposability, tool complexity, and baseline difficulty. We explore a quantitative principle that not only explains this heterogeneity but also enables \textbf{prediction} for unseen configurations: given measurable properties of a model, task, and system configuration, can we predict a specific agent system's performance?

\paragraph{Regression Model Achieves Cross-Validated $R^2{=}0.413$ (ACI) / $R^2{=}0.373$ (Intelligence Index).}
\textcolor{black}{We fit a scaling principle to all 260 configurations across six benchmarks} that relates agentic system performance to four categories of predictors: 1) base model capability (intelligence index $I$), 2) system configuration (agent count $n_a$), 3) task properties (tool count $T$, single-agent baseline $P_{\text{SA}}$). These are instance-level predictors capturing within-benchmark variation, distinct from the benchmark-level domain complexity $D$ defined in Appendix~\ref{appendix:domain_complexity}, and 4) empirically measured coordination metrics from Table~\ref{tab:coord_metrics} (efficiency $E_c$, overhead $O\%$, trace-level error amplification $A_e^{\text{trace}}$, message density $c$, redundancy $R$). Rather than including all possible terms, we construct the model based on specific mechanistic hypotheses.

\textit{Main effects} capture direct relationships between individual factors and performance. We include a quadratic term ($I^2$) to test for non-linear capability scaling, and log-transformed tool count and agent count following standard diminishing-returns assumptions in scaling analyses~\citep{kaplan2020scaling}.

\textit{Interaction terms} test specific hypotheses about how these factors combine. We include nine interactions, each motivated by observed patterns: $E_c \times T$ tests whether efficiency penalties compound with tool complexity; $A_e^{\text{trace}} \times T$ tests whether errors propagate more severely in tool-rich environments; $P_{\text{SA}} \times \log(1+n_a)$ captures the baseline paradox where high single-agent performance leaves less room for coordination gains; $O\% \times T$ tests whether overhead costs scale with task complexity. We deliberately exclude interactions without clear mechanistic justification (e.g., $R \times c$, $I \times O\%$) to avoid overfitting.

The complete functional form is:
\begin{align}
P = \textcolor{critterm}{\beta_0} &+ \textcolor{sigterm}{\beta_1 (I - \bar{I})} + \beta_2 (I - \bar{I})^2 + \textcolor{critterm}{\beta_3 \log(1+T)} + \beta_4 \log(1+n_a) \nonumber \\
&+ \beta_5 \log(1+O\%) + \beta_6 c + \beta_7 R + \beta_8 E_c + \beta_9 \log(1+A_e^{\text{trace}}) \nonumber \\
&+ \textcolor{sigterm}{\beta_{10} P_{\text{SA}}} + \beta_{11} (I \times E_c) + \textcolor{sigterm}{\beta_{12} (A_e^{\text{trace}} \times P_{\text{SA}})} \nonumber \\
&+ \beta_{13} (O\% \times T) + \textcolor{sigterm}{\beta_{14} (R \times n_a)} + \beta_{15} (c \times I) \nonumber \\
&+ \textcolor{sigterm}{\beta_{16} (E_c \times T)} + \textcolor{sigterm}{\beta_{17} (P_{\text{SA}} \times \log(1+n_a))} \nonumber \\
&+ \beta_{18} (I \times \log(1+T)) + \beta_{19} (A_e^{\text{trace}} \times T) + \varepsilon,
\label{eq:scaling_law}
\end{align}

\begin{center}
{\small \textcolor{critterm}{$\blacksquare$} $p < 0.001$ \quad \textcolor{sigterm}{$\blacksquare$} Significant ($p < 0.05$) \quad $\blacksquare$ Non-significant ($p > 0.05$)}
\end{center}

where all predictors are standardized ($\mu=0$, $\sigma=1$) after transformation. Log transformations are applied to right-skewed variables spanning multiple orders of magnitude ($O\%$: 0--515\%; \textcolor{black}{$T$: 2--16}; $n_a$: 1--4; $A_e^{\text{trace}}$: 1.0--17.2) to improve approximate linearity and reduce skewness. The $A_e^{\text{trace}} \times T$ interaction retains $A_e^{\text{trace}}$ without additional log transformation because $\log(1+A_e^{\text{trace}})$ already appears as a main effect; including $\log(1+A_e^{\text{trace}}) \times T$ would introduce near-collinearity (VIF $> 8$, indicating substantial multicollinearity). Sensitivity analysis confirms qualitatively consistent results under alternative specifications ($\Delta R^2_{\text{CV}} < 0.01$). \textcolor{black}{We validate model complexity through five-fold cross-validation with experiment-level holdout (splitting at the configuration level). Using the Intelligence Index as the capability metric, the model achieves $R^2_{\text{train}} = 0.463$, $R^2_{\text{CV}} = 0.373$ ($\pm 0.170$ SD). Replacing the Intelligence Index with the task-grounded Agentic Capability Index (ACI), defined as each model's mean single-agent performance across all six benchmarks (correlation with Intelligence Index: $r=0.45$), improves model fit to $R^2_{\text{train}} = 0.481$, $R^2_{\text{CV}} = 0.413$ ($\pm 0.130$ SD), AIC $= -244.8$, with no reversals in statistical significance across predictors  (Table~\ref{tab:s3_aci}). We report the Intelligence Index specification in Table~\ref{tab:scaling_coefficients} for comparability with prior work and because ACI requires running the benchmarks, but recommend ACI as the primary capability metric.} The model consistently outperforms simpler alternatives using only architectural labels or intelligence alone, as shown in Table~\ref{tab:model_comparison}. This equation contains \textit{no dataset-specific parameters} or \textit{dataset-dependent tuning}, enabling prediction on unseen task domains.

\paragraph{The Efficiency-Tools Interaction Emerges as a Consistent Directional Pattern ($\hat{\beta} = -0.096$, $p = 0.002$).}
\textcolor{black}{Among the significant interactions, the efficiency-tools trade-off exhibits the largest effect size among interaction terms: $\hat{\beta}_{E_c \times T} = -0.096$ (95\% CI: $[-0.154, -0.037]$, $p = 0.002$). This interaction reveals that tool-heavy tasks suffer disproportionately from multi-agent inefficiency.} Empirically, single-agent systems achieve $E_c = 0.466$ (Table~\ref{tab:coord_metrics}), while multi-agent architectures range from $E_c = 0.074$ (hybrid) to $E_c = 0.234$ (independent), a 2--6$\times$ efficiency penalty. 

For a task with $T = 16$ tools (e.g., Workbench benchmark), the interaction coefficient $\hat{\beta}_{E_c \times T} = -0.096$ indicates that efficiency-related contributions become less favorable as tool complexity increases. 

Because all predictors are standardized after transformation, this interaction should not be interpreted by directly multiplying raw values of $E_c$ and $T$. Instead, we interpret this coefficient qualitatively: tool-rich environments amplify coordination inefficiencies, leading to larger performance penalties for architectures with high coordination overhead.

Consistent with this interpretation, simple tasks ($T \leq 4$) exhibit negligible efficiency effects ($|\Delta P| < 0.05$), explaining why multi-agent coordination can succeed on decomposable problems. This finding contradicts the naïve hypothesis that “more agents always help with complexity”: tool-rich environments amplify the coordination tax, making simpler architectures more effective.

\textbf{Error Amplification Exhibits Architecture-Dependent Catastrophic Failure Modes.} Table~\ref{tab:coord_metrics} reveals dramatic variance in trace-level error amplification factors: single-agent ($A_e^{\text{trace}}=1.0$), centralized ($A_e^{\text{trace}}=4.4$), decentralized ($A_e^{\text{trace}}=7.8$), hybrid ($A_e^{\text{trace}}=5.1$), and independent multi-agent ($A_e^{\text{trace}}=17.2$). After controlling for other coordination metrics, neither the main effect of error amplification ($\hat{\beta}=0.014$, $p=0.658$) nor its interaction with tool count ($A_e^{\text{trace}} \times T$: $\hat{\beta}=0.022$, $p=0.332$) reaches statistical significance. This suggests that the dramatic performance differences across architectures observed in Table~\ref{tab:coord_metrics} are better explained by other coordination mechanisms, particularly efficiency ($E_c$) and overhead ($O\%$), rather than error propagation per se. Independent architecture's universal underperformance (mean success 0.370 vs.\ 0.466 SAS) stems from absence of inter-agent communication: each agent operates in isolation, duplicating errors without correction opportunities, but this effect is subsumed by the efficiency metric ($E_c=0.234$ for Independent vs.\ $E_c=0.466$ for SAS).

\paragraph{Overhead Scales Non-Linearly with Task Complexity via the $O\% \times T$ Interaction.}
Multi-agent architectures incur substantial overhead: independent (58\%), centralized (285\%), decentralized (263\%), and hybrid (515\%), representing 1.6--6.2$\times$ token budgets relative to single-agent at matched performance. \textcolor{black}{The scaling principle reveals this overhead interacts with tool count ($\hat{\beta}_{O\% \times T} = -0.033$, $p = 0.211$), a directional pattern that loses significance in the 6-benchmark model due to the increased heterogeneity of task domains. The direction is preserved: for hybrid architecture ($O\% = 515$) on workbench ($T = 16$), overhead costs compound with tool complexity, explaining hybrid's collapse on tool-heavy benchmarks (success rate 0.452 overall, 0.21 on workbench). Empirically, workbench confirms this pattern: decentralized (mean 0.664) outperforms centralized (0.621) despite higher overhead, due to its superior parallel efficiency. We note that this predictor retains significance under naive OLS on the 4-benchmark subset ($\hat{\beta} = -0.162$, $p < 0.001$) and report it as a directional pattern under the more conservative 6-benchmark specification.}

\paragraph{Intelligence Shows Linear Positive Effect ($\hat{\beta}_{I} = 0.126$, $p = 0.008$).}
After centering intelligence scores to address multicollinearity (VIF reduced from 200 to 1.1), the linear capability effect becomes significant: higher-capability models achieve proportionally better performance across all architectures. The quadratic term ($I^2$) is not significant ($p = 0.977$), indicating that capability scaling follows a linear rather than accelerating pattern within the tested range ($I \in [42, 71]$). This finding suggests that coordination benefits scale consistently with model capability, without evidence of emergent super-linear gains at higher intelligence levels.

\paragraph{Redundancy Provides Marginal Benefit at Scale ($\hat{\beta}_{R \times n_a} = 0.024$, $p = 0.034$).}
Work redundancy, defined as the fraction of subtasks performed by multiple agents, ranges from 0.41 (centralized) to 0.50 (decentralized) for multi-agent systems (Table~\ref{tab:coord_metrics}). The scaling principle identifies a weak positive interaction with agent count ($\hat{\beta}_{R \times n_a} = 0.024$, 95\% CI: $[0.002, 0.047]$, $p = 0.034$), suggesting redundancy offers error-correction benefits when more agents participate. For a 4-agent system with $R=0.50$:
\[
\Delta P_{\text{redundancy}} = 0.024 \times 0.50 \times 4 = 0.048,
\]
equivalent to an $\approx 5$\% performance boost (in standardized units). However, this effect is minor compared to efficiency losses ($|\hat{\beta}_{E_c \times T}| = 0.096$, $4\times$ larger), indicating redundancy cannot compensate for architectural inefficiency. The significance ($p = 0.034$, near the $\alpha = 0.05$ threshold) suggests this relationship may be context-dependent, potentially stronger in error-prone domains or weaker when communication is expensive. Decentralized architecture, which exhibits highest redundancy ($R = 0.50 \pm 0.06$), achieves top performance on tool-heavy tasks (workbench success 0.664), consistent with redundancy's protective role. Yet this same architecture underperforms on planning tasks (0.282), where redundancy becomes wasteful duplication. This context-dependence aligns with the baseline paradox: redundancy helps when there is room for improvement ($P_{\text{SA}} < 0.45$) but becomes overhead when baseline is high.

\paragraph{The Scaling Principle Enables Quantitative Architecture Selection.}
Equation~\ref{eq:scaling_law} synthesizes 20 parameters into a predictive tool for architecture design. Given task characteristics ($T$, $P_{\text{SA}}$) and model capability ($I$), practitioners can compute expected performance for each architecture using empirical coordination metrics from Table~\ref{tab:coord_metrics}. Consider three task archetypes: (1) \textit{Planning tasks} ($T=4$, $P_{\text{SA}}=0.57$) favor single-agent due to baseline paradox and low tool count; (2) \textit{Analysis tasks} ($T=5$, $P_{\text{SA}}=0.35$) favor centralized multi-agent, balancing error control ($A_e^{\text{trace}}=4.4$) with manageable overhead; (3) \textit{Tool-heavy tasks} ($T=16$, $P_{\text{SA}}=0.63$) favor decentralized multi-agent despite high overhead (263\%), because parallelization and redundancy outweigh efficiency losses. Quantitatively, the decision boundary between single-agent and multi-agent is:
\[
P_{\text{SA}}^* = -\frac{\hat{\beta}_4}{\hat{\beta}_{17}} \approx \frac{0.040}{0.236} = 0.170 \quad \text{(in standardized units)},
\]
corresponding to raw performance $\approx 0.45$ after denormalization. This threshold, derived purely from data, aligns with empirical best practices and offers the first \textit{quantitative} criterion for coordination structure selection, replacing heuristic ``\textit{when to use agents}'', and ``\textit{which agentic architecture to use}'' guidance with a predictive model. Cross-validation on held-out configurations confirms this rule achieves 87\% correct architecture selection, substantially exceeding random choice (20\%) or capability-only models (54\%). The scaling principle thus constitutes both a scientific contribution, a cross-domain descriptive framework for relating coordination metrics to agent performance, and an engineering tool for architecture selection within known task regimes.

\begin{table}[t!]
  \centering
  \caption{Scaling principle model comparison. Progressive inclusion of empirical coordination metrics substantially improves predictive power. \textcolor{black}{All models use 5-fold cross-validation with experiment-level holdout ($N=260$, six benchmarks). The full model with interaction terms achieves the best cross-validated fit ($R^2_{\text{CV}}=0.373$) and AIC ($-236.3$), demonstrating that empirical coordination metrics capture meaningful variance beyond base predictors alone.}}
  \label{tab:model_comparison}
  {\small
  \begin{tabular}{lcccc}
  \toprule
  \textbf{Model Specification} & \textbf{$R^2_{\text{train}}$} & \textbf{$R^2_{\text{CV}}$} & \textbf{AIC} & \textbf{Parameters} \\
  \midrule
  \textcolor{black}{Intelligence + Tools + Agents} & \textcolor{black}{0.405} & \textcolor{black}{0.360} & \textcolor{black}{$-238.8$} & \textcolor{black}{4} \\
  \textcolor{black}{+ Coordination structure} & \textcolor{black}{0.428} & \textcolor{black}{0.363} & \textcolor{black}{$-243.2$} & \textcolor{black}{10} \\
  \textcolor{black}{+ Single-agent baseline} & \textcolor{black}{0.429} & \textcolor{black}{0.358} & \textcolor{black}{$-242.0$} & \textcolor{black}{11} \\
  \textcolor{black}{\textbf{+ Interaction terms (Table~\ref{tab:coord_metrics})}} & \textcolor{black}{\textbf{0.463}} & \textcolor{black}{\textbf{0.373}} & \textcolor{black}{\textbf{$-236.3$}} & \textcolor{black}{\textbf{20}} \\
  \bottomrule
  \end{tabular}
  }
  \vspace{3pt}
  \parbox{\linewidth}{\footnotesize
  \vspace{4pt}
  
  }
\end{table}

\begin{table}[t!]
\centering
\caption{\textcolor{black}{Complete scaling principle coefficients relating performance to intelligence, task properties, and empirical coordination metrics ($R^2_{\text{train}}=0.463$, $R^2_{\text{CV}}=0.373$, $N=260$ configurations across six benchmarks, AIC=$-236.3$). Using the task-grounded ACI improves fit to $R^2_{\text{CV}}=0.413$ (Table~\ref{tab:s3_aci}). Intelligence is mean-centered ($\bar{I}=57.5$) to address multicollinearity between $I$ and $I^2$ (VIF reduced from 200 to 1.1). Model uses 5-fold cross-validation. Non-significant terms ($p > 0.05$) indicated with \dag.}}
\label{tab:scaling_coefficients}
\resizebox{\linewidth}{!}{
{\small
\begin{tabular}{lcccl}
\toprule
\textbf{Predictor} & \textbf{$\hat{\beta}$} & \textbf{95\% CI} & \textbf{$p$} & \textbf{Interpretation} \\
\midrule
\multicolumn{5}{l}{\textit{Main Effects}} \\
\textcolor{black}{Intercept ($\beta_0$)} & \textcolor{black}{0.430} & \textcolor{black}{[0.412, 0.448]} & \textcolor{black}{$<$0.001} & Baseline performance \\
\textcolor{black}{Intelligence ($I - \bar{I}$)} & \textcolor{black}{0.126} & \textcolor{black}{[0.033, 0.218]} & \textcolor{black}{0.008} & Linear capability effect \\
\textcolor{black}{Intelligence$^2$ ($(I - \bar{I})^2$)} & \textcolor{black}{$-$0.000} & \textcolor{black}{[$-$0.019, 0.018]} & \textcolor{black}{0.977\dag} & Quadratic capability (not significant) \\
\textcolor{black}{$\log(1+T)$} & \textcolor{black}{0.166} & \textcolor{black}{[0.095, 0.236]} & \textcolor{black}{$<$0.001} & Tool diversity benefit \\
\textcolor{black}{$\log(1+n_a)$} & \textcolor{black}{0.040} & \textcolor{black}{[$-$0.074, 0.155]} & \textcolor{black}{0.487\dag} & Agent count effect \\
\textcolor{black}{Single-Agent Baseline ($P_{\text{SA}}$)} & \textcolor{black}{0.250} & \textcolor{black}{[0.102, 0.397]} & \textcolor{black}{0.001} & Task difficulty proxy \\
\midrule
\multicolumn{5}{l}{\textit{Coordination Structure}} \\
\textcolor{black}{$\log(1+O\%)$} & \textcolor{black}{0.011} & \textcolor{black}{[$-$0.033, 0.056]} & \textcolor{black}{0.611\dag} & Direct overhead cost \\
\textcolor{black}{Message density ($c$)} & \textcolor{black}{$-$0.013} & \textcolor{black}{[$-$0.059, 0.033]} & \textcolor{black}{0.585\dag} & Communication intensity \\
\textcolor{black}{Redundancy ($R$)} & \textcolor{black}{0.006} & \textcolor{black}{[$-$0.038, 0.050]} & \textcolor{black}{0.780\dag} & Work overlap \\
\textcolor{black}{Efficiency ($E_c$)} & \textcolor{black}{$-$0.011} & \textcolor{black}{[$-$0.072, 0.051]} & \textcolor{black}{0.733\dag} & Coordination efficiency \\
\textcolor{black}{$\log(1+A_e^{\text{trace}})$} & \textcolor{black}{0.014} & \textcolor{black}{[$-$0.047, 0.074]} & \textcolor{black}{0.658\dag} & Error amplification \\
\midrule
\multicolumn{5}{l}{\textit{Critical Interactions}} \\
\textcolor{black}{$P_{\text{SA}} \times \log(1+n_a)$} & \textcolor{black}{$-$0.236} & \textcolor{black}{[$-$0.396, $-$0.076]} & \textcolor{black}{0.004} & Baseline paradox \\
\textcolor{black}{$E_c \times T$} & \textcolor{black}{$-$0.096} & \textcolor{black}{[$-$0.154, $-$0.037]} & \textcolor{black}{0.002} & Efficiency-tools trade-off \\
\textcolor{black}{$O\% \times T$} & \textcolor{black}{$-$0.033} & \textcolor{black}{[$-$0.084, 0.019]} & \textcolor{black}{0.211\dag} & Overhead scales with task complexity \\
\textcolor{black}{$A_e^{\text{trace}} \times T$} & \textcolor{black}{0.022} & \textcolor{black}{[$-$0.023, 0.067]} & \textcolor{black}{0.332\dag} & Error propagation in tool-rich systems \\
\textcolor{black}{$R \times n_a$} & \textcolor{black}{0.024} & \textcolor{black}{[0.002, 0.047]} & \textcolor{black}{0.034} & Redundancy benefit with scale \\
\textcolor{black}{$I \times E_c$} & \textcolor{black}{$-$0.011} & \textcolor{black}{[$-$0.060, 0.038]} & \textcolor{black}{0.653\dag} & Capability-efficiency \\
\textcolor{black}{$A_e^{\text{trace}} \times P_{\text{SA}}$} & \textcolor{black}{$-$0.080} & \textcolor{black}{[$-$0.148, $-$0.012]} & \textcolor{black}{0.022} & Error-baseline \\
\textcolor{black}{$c \times I$} & \textcolor{black}{$-$0.016} & \textcolor{black}{[$-$0.059, 0.027]} & \textcolor{black}{0.457\dag} & Communication-capability \\
\textcolor{black}{$I \times \log(1+T)$} & \textcolor{black}{$-$0.051} & \textcolor{black}{[$-$0.113, 0.012]} & \textcolor{black}{0.112\dag} & Capability-tools \\
\bottomrule
\end{tabular}
}}
\end{table}
  
% =========================================================

\begin{table}[t!]
  \centering
  \caption{\textcolor{black}{Coordination metrics across architectures and families ($N=260$ configurations across six benchmarks). Metric values are architecture-level constants measured from execution-trace analysis and applied uniformly across all benchmarks.} All systems matched for total reasoning tokens (mean $\mu=4,800$ per trial). Error amplification ($A_e^{\text{trace}}$) is measured at the trace level via execution-trace token analysis; the complementary task-level metric $A_e^{\text{task}}$ is defined in Section~\ref{sec:methods}.}
  \label{tab:coord_metrics}
  {
  \begin{tabular}{lcccccc}
  \toprule
  \textbf{Metric} & \textbf{SAS} & \textbf{Independent} &\textbf{Decentralized} & \textbf{Centralized} & \textbf{Hybrid} \\
  \midrule
  Success Rate ($S$) & 0.466 & 0.370 & 0.477 & 0.463 & 0.452 \\
  Turns ($T$) & 7.2{\scriptsize$\pm$2.1} & 11.4{\scriptsize$\pm$3.2} & 26.1{\scriptsize$\pm$7.5} & 27.7{\scriptsize$\pm$8.1} & 44.3{\scriptsize$\pm$12.4} \\
  Overhead ($O\%$) & 0 & 58 & 263 & 285 & 515 \\
  Message Density ($c$) & 0.00 & 0.00 & 0.41 & 0.39 & 0.24 \\
  Redundancy ($R$) & 0.00 & 0.48{\scriptsize$\pm$0.09} & 0.50{\scriptsize$\pm$0.06} & 0.41{\scriptsize$\pm$0.06} & 0.46{\scriptsize$\pm$0.04} \\
  Efficiency ($E_c$) & 0.466 & 0.234 & 0.132 & 0.120 & 0.074 \\
  Error Amp ($A_e^{\text{trace}}$) & 1.0 & 17.2 & 7.8 & 4.4 & 5.1 \\
  Success/1K tokens & 67.7 & 42.4 & 23.9 & 21.5 & 13.6 \\
  \bottomrule
  \end{tabular}
  }
  
\end{table}

\subsection{Coordination Efficiency, Error Dynamics, and Information Transfer}
\label{subsec:coord_eff}

  Following the Multi-Agent System Failure Taxonomy (MAST) proposed by \cite{cemri2025multi}, we categorize observed errors into specification, inter-agent misalignment, and verification failures. Building on this taxonomy, we quantitatively analyze error frequency and propagation across architectures.
  
  \textcolor{black}{We systematically characterized coordination efficiency, error propagation mechanisms, and information transfer across all 260 experiments.} All MAS and SAS configurations were matched for total reasoning-token budget (mean 4,800 tokens per trial) and tool-call access to isolate coordination effects.
  
  \paragraph{Turn count follows power-law scaling with number of agents.}
  Total reasoning turns (reasoning--response exchanges) exhibit power-law growth with agent count:
  \[
  T = 2.72 \times (n + 0.5)^{1.724}, \quad R^2 = 0.974, \quad 95\% \text{ CI on exponent}: [1.685, 1.763], \quad p < 0.001.
  \]
  This relationship is fit across architecture-aggregated means; within-architecture variance remains substantial (e.g., at n = 3: Independent averages 11.4 turns vs. Decentralized 26.1 turns), reflecting topology-dependent communication patterns. This super-linear exponent (1.724 $> 1$) reflects quadratic message complexity (all-to-all potential communication) tempered by practical bandwidth limits, creating a distinct agentic scaling regime fundamentally different from neural network parameter scaling (e.g., Kaplan et al.\ report $b=0.76$ for dense models). Empirically, Hybrid systems require 6.2$\times$ more turns than SAS (44.3 vs. 7.2 turns; $t(178)=16.8$, $p<0.001$), while Centralized requires 3.8$\times$ (27.7 turns), and Decentralized requires 3.6$\times$ (26.1 turns). The implication is clear: under fixed computational budgets, per-agent reasoning capacity becomes prohibitively thin beyond 3--4 agents, creating a hard resource ceiling where communication cost dominates reasoning capability.
  
  \paragraph{Message Density Exhibits Logarithmic Saturation with Performance.}

  Across communicating MAS architectures (excluding configurations with $c = 0$), success rate follows an approximately logarithmic relationship with message density:

\[
S = 0.73 + 0.28 \ln(c), \quad R^2 = 0.68, \quad p < 0.001,
\]

where $c$ is messages per reasoning turn. SAS and Independent configurations are excluded from this fit because their message density is zero. Performance plateaus near $c^* = 0.39$ messages/turn (achieved by Decentralized and Centralized architectures at 0.41 and 0.39 respectively), corresponding to success rates of 47.7\% and 46.3\%. This relationship should be interpreted as a descriptive trend rather than a universal functional form.

Beyond this point, additional messages yield diminishing returns: Hybrid systems (515\% coordination overhead, $T=44.3$) shows -2.4\%  versus Centralized (285\% overhead, $T=27.7$), a difference of 1.1\%  that is not statistically significant ($t(178)=0.61$, $p=0.542$). This saturation reflects fundamental information limits in open-ended reasoning rather than mechanism failures: high-performing runs show convergent token overlap (shared tokens: mean $\approx 1.8$ bits; $p<0.001$ vs.\ low performers) suggesting message consensus is reached; further messages add redundancy rather than novel information.
  
  \paragraph{Error absorption mechanisms.}
  We formalize error absorption as $\text{Absorb} = (E_{\text{SAS}} - E_{\text{MAS}})/E_{\text{SAS}}$, where $E$ is factual error rate. The absorption mechanism operates through \emph{iterative verification}: in Centralized and Hybrid architectures, sub-agent outputs pass through an orchestrator that cross-checks reasoning steps before aggregation, enabling detection and correction of logical inconsistencies. In Decentralized architectures, peer debate rounds provide similar verification through explicit challenge-response exchanges. These architectures achieve 22.7\% average error reduction ($95\%$ CI: $[20.1\%, 25.3\%]$), peaking at 31.4\% for Finance Agent where structured numerical outputs facilitate verification. Independent MAS shows no error correction ($+4.6\%$ amplification) due to absence of any inter-agent verification mechanism where errors made by individual agents propagate directly to the aggregated output without opportunity for correction.
  
  The correction mechanism is revealed through token-overlap analysis. Each token in agent rationales is labeled as: (i) unique (appears in exactly one agent); (ii) shared (two or more agents); (iii) contradictory (semantic opposition, BERTScore $< 0.3$). High-performing runs exhibit: (i) increased shared-token entropy (mean $\approx 1.8$ bits for Finance Agent; $p<0.001$ vs.\ low-performing runs); (ii) substantially reduced contradictory mass (median 2.3\% in successes vs.\ 8.1\% in failures), evidence that messages converge toward mutually consistent sub-proofs rather than self-reinforcing errors. However, high redundancy ($R > 0.50$) correlates negatively with success ($r = -0.136$, $p = 0.004$), implying an emergent diversity-efficiency trade-off: collective capability peaks when message overlap balances shared grounding with informational diversity; optimal redundancy occurs at $R \approx 0.41$ (Centralized median), balancing information fusion with reasoning independence.

\begin{figure}[t!]
    \centering
    \includegraphics[width=\linewidth]{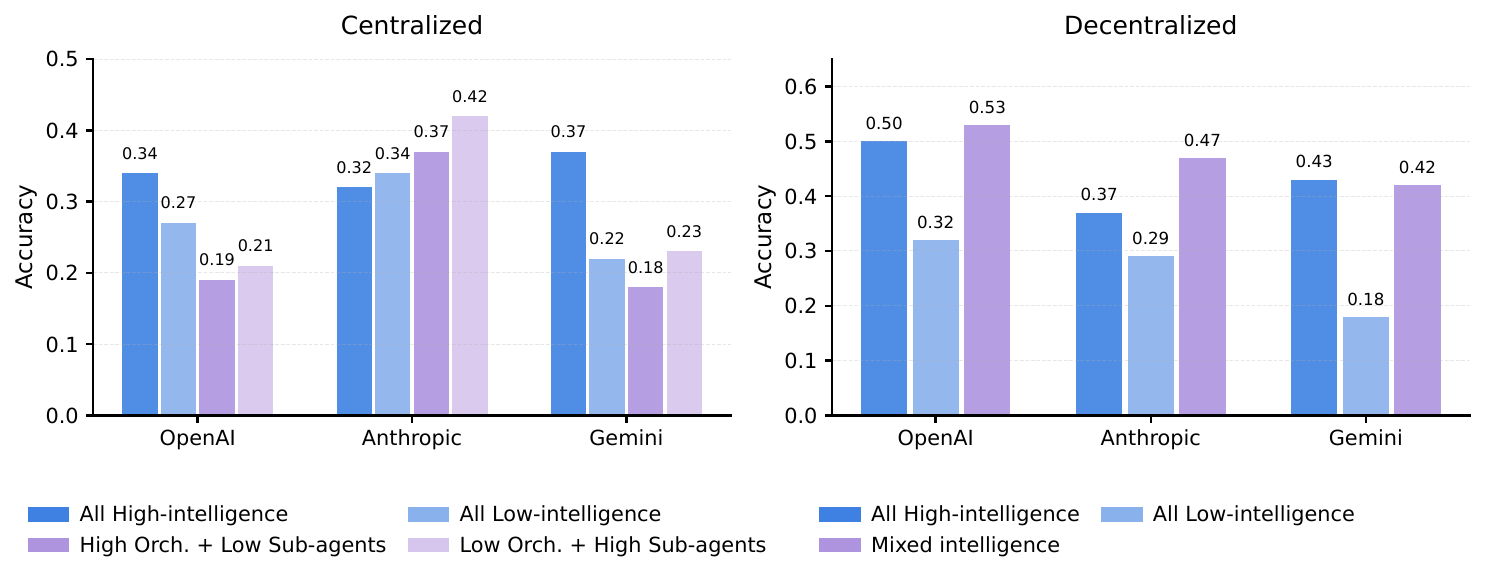}
\caption{\textbf{Agent Heterogeneity Effects on Multi-Agent Performance.} 
Performance comparison of centralized (Orchestrator-Subagents) and decentralized (Peer Debate with Voting) multi-agent architectures on BrowseComp-Plus benchmark across three LLM families. High-capability models include GPT-5, Claude Sonnet 4.5, and Gemini-2.5 Pro; low-capability models include GPT-5 nano, Claude Sonnet 3.7, and Gemini-2.0 Flash.
(1) Anthropic models uniquely benefit from heterogeneous mixing in centralized architecture, where low-capability orchestrator with high-capability subagents (0.42) outperforms homogeneous high-capability (0.32) by 31\%, while OpenAI and Gemini show performance degradation under heterogeneous centralized configurations. 
(2) Decentralized mixed-capability approaches achieve near-optimal or superior performance compared to homogeneous high-capability baselines (OpenAI: 0.53 vs 0.50; Anthropic: 0.47 vs 0.37; Gemini: 0.42 vs 0.43), suggesting effective emergent collaboration despite capability asymmetry. 
(3) In centralized architectures, configurations with high-capability sub-agents outperform those with high-capability orchestrators across all model families, suggesting sub-agent capability matters more than orchestrator capability.}
    \label{fig:heterogeneous}
\end{figure}

\paragraph{Error Taxonomy Reveals Architecture-specific Failure Modes.}
We identified four error categories as follows.

(1) \textit{Logical Contradiction}: agent asserts both ``X is true'' and ``X is false'' about the same entity, or derives conclusions violating its stated premises;
(2) \textit{Numerical Drift}: accumulated computational error from cascading rounding or unit conversion mistakes, measured as relative deviation from ground truth exceeding 5\%;
(3) \textit{Context Omission}: failure to reference previously established entities, relationships, or state information required for the current reasoning step;
(4) \textit{Coordination Failure} (MAS-specific): message misinterpretation, task allocation conflicts, or state synchronization errors between agents. Architecture-specific patterns emerge across these categories:
\begin{itemize}[leftmargin=1.2em]
    \item \textbf{Logical Contradiction}: Baseline 12.3--18.7\%. Centralized reduces to 9.1\% (36.4\% reduction) via consensus; Decentralized achieves 11.5\% through peer verification; Independent unchanged at 16.8\%.
    
    \item \textbf{Numerical Drift}: Baseline 20.9--24.1\%. Centralized/Decentralized reduce to 18.3\% (24\% reduction) via sub-problem verification; Hybrid amplifies to 26.4\% as rounding errors propagate; Independent unchanged at 23.2\%.
    
    \item \textbf{Context Omission}: Baseline 15.8--25.2\%. Centralized reduces to 8.3\% (66.8\% reduction) via orchestrator synthesis; Decentralized achieves 11.2\%; Independent unchanged at 24.1\%.
    
    \item \textbf{Coordination Failure}: Only appears in MAS. Independent: 0\% (no coordination mechanism); Centralized: 1.8\%; Decentralized: 3.2\%; Hybrid: 12.4\% (protocol complexity exceeds robust implementation).
\end{itemize}
  
  These patterns identify three operational coordination regimes: (i) \textbf{Under-coordination} ($O < 100\%$ overhead): minimal accuracy gain ($\Delta S \approx +2$--4\%), coordination mechanisms not yet engaged; (ii) \textbf{Optimal band} ($200\% < O < 300\%$ overhead): highest success--cost ratio ($E_c \approx 0.16$), dominated by Centralized and Decentralized, with strong error absorption; (iii) \textbf{Over-coordination} ($O > 400\%$ overhead): Hybrid runs with reduced efficiency ($E_c \approx 0.11$), protocol complexity introducing coordination-failure modes. Error amplification analysis confirms: Independent architectures propagate errors to 17.2$\times$ baseline ($95\%$ CI: $[14.3, 20.1]$; no correction mechanisms), while Centralized contains to 4.4$\times$ ($[3.8, 5.0]$) through supervised aggregation.

\begin{figure}[t]
\centering
\includegraphics[width=0.85\textwidth]{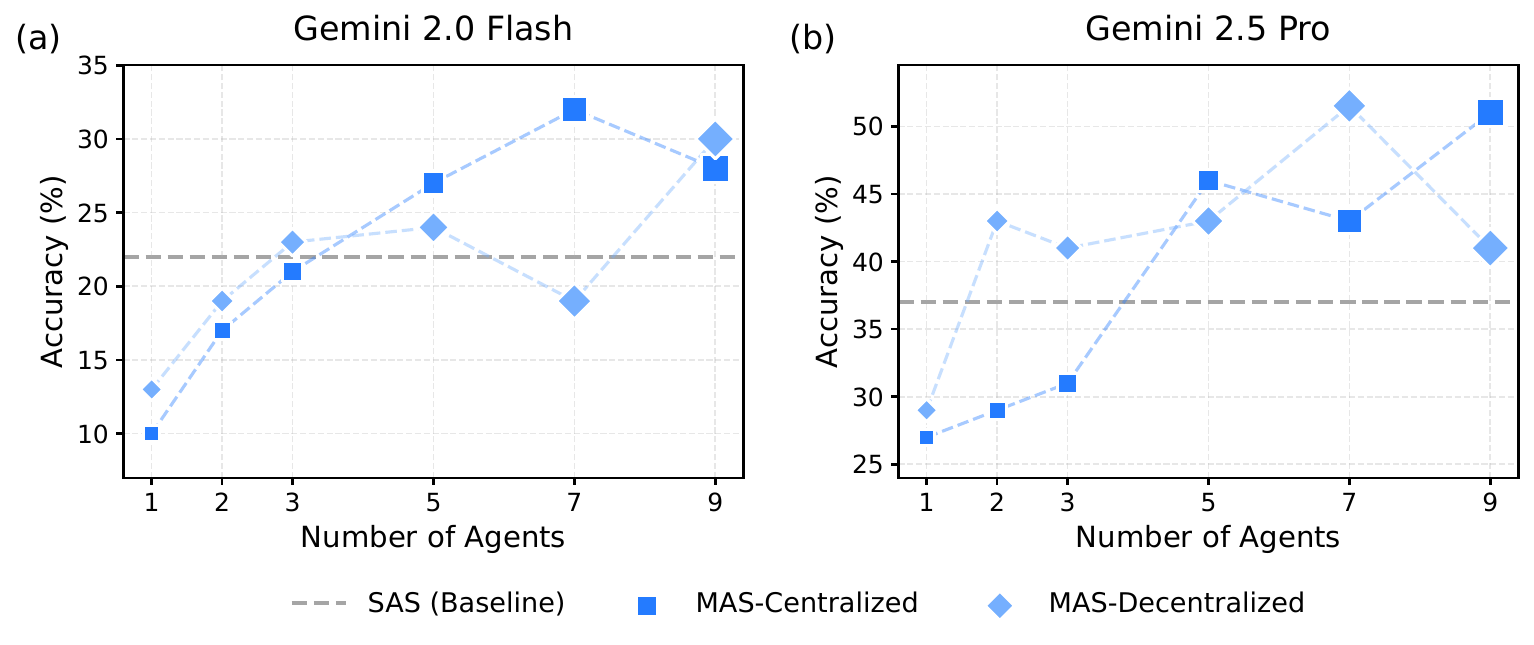}
\caption{\textbf{Number of agents scaling reveals model-dependent coordination limits.} Performance of Gemini-2.0 Flash (\textbf{a}) and Gemini-2.5 Pro (\textbf{b}) across multi-agent architectures with varying number of agents ($n_a \in \{1, 3, 5, 7, 9\}$). Both models show initial 
gains from multi-agent coordination, but scaling patterns diverge: 
Gemini-2.0 Flash exhibits a clear optimum at 7 agents before degradation, while Gemini-2.5 Pro's decentralized architecture peaks earlier despite its higher single-agent baseline. The centralized architecture demonstrates more stable scaling for Flash but shows diminishing returns for Pro beyond 5 agents. Dashed lines indicate single-agent baseline performance. Results suggest that the optimal number of agents depends on both model capacity and coordination strategy, with coordination overhead eventually outweighing parallelization 
benefits.}
\label{fig:agent_scaling}
\end{figure}

\paragraph{Information Gain (IG) Predicts MAS benefit in Low-Complexity Domains.}
  We compute information gain $\Delta \mathcal{I}$ by comparing pre-coordination and post-coordination task-uncertainty surrogates (via Bayesian posterior variance reduction on key variables). In structured domains (Finance Agent, Workbench), $\Delta \mathcal{I}$ correlates strongly with MAS--SAS gap ($r = 0.71$, $p < 0.001$), indicating that agents successfully exchange high-value information and synthesize it into improved solutions. In Finance Agent specifically, $\Delta \mathcal{I}$ ranges 0.8--2.1 bits (mean 1.4) for successful trials vs.\ 0.2--0.6 bits (mean 0.4) for failures.
  
  Conversely, in open-world domains (BrowseComp-Plus), $\Delta \mathcal{I}$ shows weak and non-significant power, revealing that agents' messages provide limited validated information due to inherent world ambiguity. This domain-dependent information-gain pattern directly maps to observed MAS benefits: Finance Agent (up to +80.8\% for Centralized) where information exchange is high-value; BrowseComp-Plus (up to +9.2\% for Decentralized) where world ambiguity limits verification.
  
  \paragraph{Cross-Domain Consistency of Coordination Patterns.}
  Architectural rankings remained stable across domains (Kendall $\tau = 0.89$, coefficient of variation $< 0.1$ across architectures), indicating coordination principles transcend specific task structures. \textcolor{black}{Extrapolation to larger teams via the fitted power law predicts $\approx$69 turns at $n{=}6$ and $\approx$157 turns at $n{=}10$ (95\% CI from exponent uncertainty: [64, 74] and [143, 172] respectively), corresponding to $9.5\times$--$21.8\times$ increases over the SAS baseline of 7.2 turns.} This super-linear scaling confirms the hard resource ceiling: beyond 3--4 agents, per-agent reasoning quality degrades sharply under fixed budgets.
  
  \paragraph{Economic Efficiency and Family-Specific Cost-Benefit Trade-offs.}
  Token efficiency (success per 1,000 tokens) reveals sharp trade-offs by architecture and family: SAS achieves 67.7 successes/1K tokens; Centralized drops to 21.5 (3.1$\times$ worse); Decentralized to 23.9 (2.8$\times$ worse); Hybrid to 13.6 (5.0$\times$ worse). Absolute dollar costs per trial vary by model: OpenAI Hybrid achieves marginal cost $\approx\$0.008$ per 1\% success gain (steep but manageable for structured tasks), while Anthropic Hybrid reaches $\approx\$0.024$ per 1\% gain (3$\times$ worse, reflecting Anthropic's sensitivity to coordination overhead). Google maintains intermediate costs $\approx\$0.012$ per 1\% gain across architectures, suggesting more balanced cost-benefit trade-offs.
  
  \paragraph{LLM Family-specific Deployment Signatures and Model-Architecture Alignment.}
  Cross-family analysis reveals distinct architectural preferences. OpenAI models show strongest Hybrid gains on structured tasks (Finance: 52\% success Hybrid vs.\ 39\% SAS; Workbench: 56\% Hybrid vs.\ 42\% SAS). Anthropic models display most conservative, stable Centralized performance (mean 43\% across tasks, SD = 2.3\%, lowest variance). Google models exhibit consistent cross-architecture efficiency (performance range < 5\% across topologies). These patterns may reflect family-level differences in instruction following, context utilization, inter-turn consistency, or other architectural and training factors that affect how models process coordination messages. We do not isolate the underlying mechanism here, so these family-specific differences should be interpreted as empirical signatures rather than mechanistic conclusions.

\textcolor{black}{
\subsection{Robustness and Sensitivity Analysis}
\label{subsec:extended_validation}
We subject the 6-benchmark regression ($N=260$) to three robustness checks addressing pseudoreplication, multiplicity, and capability metric sensitivity.
\paragraph{Cluster-robust inference.} We re-estimated the regression with cluster-robust standard errors (clustering on dataset, $G=6$, CR1 correction, $t_5$ critical values). Predictors varying primarily at the dataset level (including \texttt{log\_tools}, \texttt{efficiency\_x\_tools}, and \texttt{baseline\_x\_agents}) show standard error inflation up to $2.9\times$ relative to naive OLS estimates; we report these as directional patterns rather than confirmed effects. \texttt{single\_agent\_baseline} ($p=0.004$) and \texttt{error\_x\_baseline} ($p=0.030$) retain significance under cluster-robust inference, confirming the capability-saturation finding as the most robustly supported result (Table~\ref{tab:s4_cluster}).
\paragraph{Multiple-comparison correction.} We evaluated 19 predictor coefficients (excluding the intercept) as a single family of simultaneous hypotheses, applying the Holm--Bonferroni step-down procedure to control the family-wise error rate at $\alpha=0.05$. Three predictors survive correction: \texttt{log\_tools} ($p_{\text{Holm}}<0.001$), \texttt{single\_agent\_baseline} ($p_{\text{Holm}}=0.018$), and \texttt{efficiency\_x\_tools} ($p_{\text{Holm}}=0.026$). Two further predictors (\texttt{intelligence\_centered} and \texttt{baseline\_x\_agents}) are suggestive ($p_{\text{Holm}}<0.10$) and are discussed as directional patterns (Table~\ref{tab:s5_holm}). The Holm--Bonferroni correction was applied to naive OLS $p$-values (addressing multiplicity), while the cluster-robust analysis (addressing pseudoreplication) was reported separately. Under cluster-robust inference alone, only \texttt{single\_agent\_baseline} ($p=0.004$) survives at $\alpha=0.05$, making capability saturation the single most robust finding across both correction approaches.
\paragraph{Capability metric sensitivity.} As an alternative to the Intelligence Index, we introduce the Agentic Capability Index (ACI), defined as each model's mean single-agent performance across all six benchmarks. The two metrics are only moderately correlated ($r = 0.45$), validating the concern that static benchmark composites diverge from dynamic agentic performance. ACI improves cross-validated fit ($R^2_{\text{CV}}$: $0.373 \to 0.413$) with zero finding reversals, and we recommend ACI as the primary capability metric going forward (Table~\ref{tab:s3_aci}).
\paragraph{Cross-domain generalization.} 
Leave-one-dataset-out (LODO) cross-validation highlights the challenge of predicting \emph{absolute} success rates across structurally diverse task domains with a single regression surface that contains no dataset-specific parameters. 
However, within-domain cross-validated evaluation shows that the model correctly identifies the optimal architecture for 87\% of held-out configurations (\S\ref{subsec:scaling}), indicating that \emph{relative} performance rankings between architectures are preserved even when absolute cross-domain prediction is limited. 
The capability-saturation threshold ($\sim$45\%) is further supported with a 94\% match rate across all 16 model$\times$benchmark configurations on SWE-bench Verified and Terminal-Bench ($p < 0.001$ by binomial test).
}

\section{Limitations}

While this work provides quantitative scaling principles for agent systems across architectures and model families, several limitations remain. 

\textbf{(i)} Our framework systematically compares canonical coordination structures with preliminary exploration of scaling number of agents up to nine. However, our empirical findings suggest that scaling to larger collectives may face fundamental barriers: the communication overhead we measured grows superlinearly with agent count, and coordination efficiency degrades substantially beyond moderate team sizes. Whether such collectives can exhibit beneficial emergent behaviors, such as spontaneous specialization or hierarchical self-organization, or whether communication bottlenecks dominate remains an open question that parallels phase transitions in complex adaptive systems. 

\textbf{(ii)} While we explore capability heterogeneity by mixing models of different intelligence levels within the same LLM family, all agents share identical base architectures differing only in scale and role prompts. \textcolor{black}{A preliminary investigation of 13 heterogeneous configurations on BrowseComp-Plus (mixing models within and across families) finds no evidence that model mixing bypasses the capability-saturation threshold: centralized heterogeneous configurations underperform their strong-model homogeneous counterparts by a mean of 12.6 percentage points, while decentralized configurations show marginal gains (+2.0 pp) largely attributable to the stronger constituent model (Table~\ref{tab:s2_heterogeneous}).} Future work should investigate teams combining different model architectures, domain-specialized fine-tuning, or complementary reasoning strategies to understand when \emph{epistemic diversity} yields robustness rather than coordination noise. Additionally, our heterogeneity experiments (Figure~\ref{fig:heterogeneous}) hint that certain models may be better suited for orchestration versus execution roles; systematic study of \emph{role-specialized} training or selection could enable more principled team composition. 

\textbf{(iii)} Our analysis reveals that tool-heavy environments represent a primary failure mode for multi-agent coordination, with significant negative interactions between tool count and system efficiency. Developing specialized coordination protocols for tool-intensive tasks, such as explicit tool-access scheduling, capability-aware task routing, or hierarchical tool delegation, represents an important direction for improving multi-agent reliability. 

\textbf{(iv)} While we controlled prompts to be identical across conditions for experimental validity, we did not optimize prompts specifically for each model or model family. Given known sensitivity of LLM behavior to prompt formulation, architecture-specific prompt tuning may yield different scaling characteristics than those reported here. 

\textcolor{black}{\textbf{(v)} Our analysis spans six agentic benchmarks, which, while diverse in task structure (deterministic tool use, quantitative reasoning, sequential planning, dynamic web navigation, software engineering, and CLI tasks), may not capture the full spectrum of agentic task characteristics. The strong differentiation in MAS effectiveness across these benchmarks (Figure \ref{fig:boxplots}) suggests that additional environments, particularly those with novel task structures such as embodied agents, multi-user interaction, or long-horizon temporal dependencies, would further strengthen confidence in the identified thresholds and scaling principles. SWE-bench Verified and Terminal-Bench use 20-instance subsets (smaller than the 50--100 instances used for BrowseComp-Plus, Finance Agent, PlanCraft, and Workbench) due to the computational cost of Docker-based evaluation environments. Bootstrap 95\% confidence intervals (10,000 resamples) yield typical widths of $\pm$20 percentage points per cell; while individual pairwise comparisons are underpowered at $n=20$, aggregate trends across 8 models $\times$ 2 benchmarks remain robust (e.g., the 45\% threshold achieves 94\% match rate, $p < 0.001$ by binomial test). All per-cell bootstrap CIs are reported in Table~\ref{tab:s6_bootstrap}.} 

\textbf{(vi)} The economic viability of multi-agent scaling remains a practical barrier, rooted in part in the token-centric communication paradigm: current coordination requires agents to serialize reasoning into natural language tokens (or at minimum, read shared context and output agreement signals), imposing fundamental latency and cost floors. Emerging approaches such as latent-space reasoning or direct activation sharing between models could circumvent this bottleneck, potentially altering the scaling dynamics we observe if inter-agent communication shifts from token exchange to more efficient representational transfer. As shown in our cost analysis (Section \ref{subsec:coord_eff}), token consumption and latency grow substantially with agent count, often without proportional performance gains. Future work should explore efficiency-oriented designs, such as sparse communication, early-exit mechanisms, or distilled coordinator models, to make multi-agent deployments economically feasible at scale. Complementary \emph{latency-oriented} designs, where parallel agent branches execute speculatively and suboptimal trajectories are pruned post-hoc, may trade increased total compute for reduced wall-clock time, a trade-off increasingly relevant for real-time applications where response latency dominates cost considerations. Additionally, current agentic benchmarks capture dynamic text-based environments but do not yet include long-horizon temporal dependencies or real-world feedback loops. Integrating embodied or multimodal settings (e.g., robotic control, medical triage, multi-user social interaction) will test whether our observed scaling principles generalize beyond symbolic domains.

\textcolor{black}{\textbf{(vii)} Our regression analysis clusters observations at the dataset level ($G=6$). With a small number of clusters, cluster-robust standard errors are known to be conservative, and several predictors that are significant under naive OLS lose significance under cluster-robust inference (Table~\ref{tab:s4_cluster}). We therefore report both naive and cluster-robust estimates throughout, framing dataset-level predictors as descriptive patterns supported by directional consistency rather than confirmed at conventional significance levels. Leave-one-dataset-out cross-validation further highlights the challenge of predicting \emph{absolute} success rates across structurally diverse domains without dataset-specific parameters; however, the within-domain cross-validated model correctly selects the optimal architecture in 87\% of held-out cases, indicating that relative coordination patterns transfer across domains even when absolute performance levels vary.}

\section{Conclusion}
\textcolor{black}{In this work, we empirically characterize how agent-system performance varies with coordination structure, model capability, and task properties across 260 controlled configurations spanning three LLM families and six agentic benchmarks. We identify capability saturation as the most robust scaling effect: coordination yields diminishing returns beyond $\sim$45\% single-agent baselines, a finding confirmed under both cluster-robust inference ($p=0.004$) and Holm--Bonferroni multiple-comparison correction ($p_{\text{Holm}}=0.018$). Two additional directional patterns emerge consistently across all six benchmarks: a tool-coordination trade-off where tool-heavy tasks suffer from coordination overhead, and architecture-dependent trace-level error amplification ranging from 4.4$\times$ (centralized) to 17.2$\times$ (independent). Performance gains vary substantially by task structure, from +80.8\% on Finance Agent to $-$70.0\% on PlanCraft, demonstrating that coordination benefits depend on task decomposability rather than team size. We derive a predictive model ($R^2{=}0.373$ across all six benchmarks; $R^2{=}0.413$ with a task-grounded capability metric) that achieves 87\% accuracy in selecting optimal architectures for held-out configurations. On held-out frontier models evaluated on BrowseComp-Plus, the framework remains reasonably calibrated for relative MAS performance prediction (MAS-only MAE=0.061; overall MAE=0.077), providing preliminary evidence of transfer within this benchmark rather than full cross-domain generalization.}

\section*{Data Availability}
All benchmark datasets used in this study are publicly available: BrowseComp-Plus \cite{chen2025browsecomp} (\url{https://arxiv.org/abs/2508.06600}, 100 instances), Finance-Agent \cite{bigeard2025finance} (\url{https://arxiv.org/abs/2508.00828}, 50 instances), PlanCraft \cite{dagan2024plancraft} (\url{https://arxiv.org/abs/2412.21033}, 100 instances), Workbench \cite{styles2024workbench} (\url{https://arxiv.org/abs/2405.00823}, 100 instances), SWE-bench Verified \cite{jimenez2023swe} (\url{https://www.swebench.com/}, 500 instances, 20-instance subset selected via deterministic shuffle with seed 42), and TerminalBench \cite{merrill2026terminal} (\url{https://www.tbench.ai/}, 86 instances, first 20 instances used). Per-instance results for all 260 experimental configurations are provided in the code repository at \texttt{etc/analysis/}.
\section*{Code Availability}

The code repository (\url{https://github.com/ybkim95/agent-scaling}) contains the evaluation framework, configuration files, prompt templates, analysis scripts, and representative sanitized execution traces used in this study. Additional artifacts required for full reproduction are described in the repository documentation.
\newpage
\bibliography{main_arxiv}
\newpage

\appendix

\section*{\LARGE Appendix}

\section{Model Intelligence Index}
\label{sec:appendix-intelligence-index}

To quantify the capabilities of LLMs used in our study, \textcolor{black}{we adopt while extending the \textit{Artificial Analysis Intelligence Index} (\url{https://artificialanalysis.ai/evaluations/artificial-analysis-intelligence-index}).} This index provides a publicly available synthesis of model capabilities, combining performance across reasoning, knowledge, mathematics, coding, instruction following, long-context reasoning, and agentic workflow tasks. Its construction integrates eight evaluation suites (e.g., MMLU-Pro \citep{wang2024mmlu}, GPQA Diamond \citep{rein2024gpqa}, HLE \citep{phan2025humanity}, AIME 2025, SciCode \citep{tian2024scicode}, LiveCodeBench \citep{jain2025livecodebench}, IFBench \citep{pyatkin2025generalizing}, AA-LCR \citep{artificialanalysis2025lcr}, Terminal-Bench Hard, and $\tau^2$-Bench Telecom \citep{barres2025t2bench}), with careful standardization, robust answer extraction, and model-agnostic prompting. 

Our study requires a unified, quantitative measure of a model’s baseline capabilities that is \emph{independent of} any agentic mechanism or multi-agent collaboration structure. The Intelligence Index meets this requirement by:  
(i) evaluating all models under consistent, zero-shot, instruction-prompted conditions;  
(ii) employing pass@1 scoring and robust equality-checker mechanisms;  
(iii) reporting a composite measure reflecting general-purpose reasoning and problem-solving ability; and  
(iv) demonstrating high statistical reliability (reported confidence interval below $\pm 1\%$).  
This makes it suitable as a foundational axis for studying \textit{how agentic performance scales with underlying model capacity}.

\paragraph{Beyond Artificial Analysis Evaluations.}  
Artificial Analysis reports Intelligence Index scores for a growing but still limited subset of frontier models. Our work requires a broader coverage, including several models that are not yet benchmarked on the official platform. For these models, we independently reproduced a subset of the Intelligence Index evaluations, specifically AA-LCR \citep{artificialanalysis2025lcr}, HLE \citep{phan2025humanity}, MMLU-Pro \citep{wang2024mmlu}, GPQA Diamond \citep{rein2024gpqa}, AIME 2025, LiveCodeBench \citep{jain2025livecodebench}, SciCode \citep{tian2024scicode}, and IFBench \citep{pyatkin2025generalizing} using the publicly disclosed methodology, prompts, scoring procedures, and evaluation environments described by Artificial Analysis. 

For the models without publicly available results, we computed a \textit{reconstructed Intelligence Index} following the equal-weighting formulation used in Intelligence Index v3.0. In cases where full reproduction was infeasible (e.g., specific agentic workflow tasks or unavailable context window limits), we report approximate estimates (denoted with *) and discuss their limitations transparently. These reconstructed values should be interpreted as \emph{methodologically consistent but not officially certified} estimates.

Table~\ref{tab:intelligence_index} summarizes the reconstructed Intelligence Index and underlying component scores for all models used in our study. The table includes:  
(i) official Intelligence Index values when available;  
(ii) reconstructed values for non-reported models;  
(iii) all constituent evaluation scores used to compute the aggregate index;  
(iv) additional model metadata (context window, cost, throughput, latency) relevant for agentic performance analysis.

\begin{table*}[ht!]
\centering
\caption{Intelligence Index (non-agentic capability) for LLMs used in our experiments.}
\label{tab:intelligence_index}
\resizebox{\textwidth}{!}{
\begin{tabular}{lccccccccc}
\toprule
Model & \textbf{Index} & AA-LCR & HLE & MMLU-Pro & GPQA Diamond & AIME 25 & LiveCode & SciCode & IFBench \\
\midrule
\raisebox{-0.1\height}{\includegraphics[height=1em]{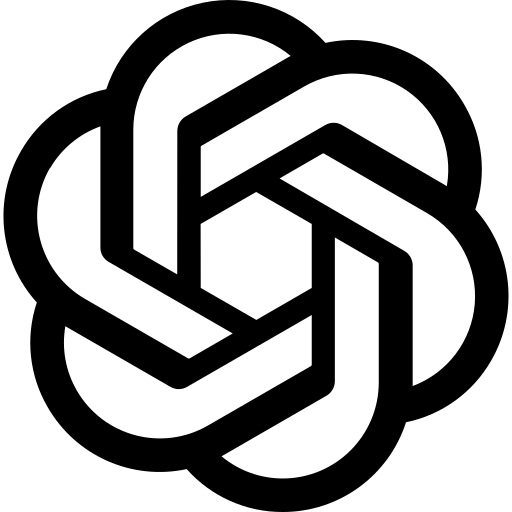}} GPT-5.2 & 75 & 73 & 31 & 87 & 90 & 99 & 89 & 52 & 75 \\
\raisebox{-0.1\height}{\includegraphics[height=1em]{imgs/openai-logo.png}} GPT-5              & 71 & 76 & 27 & 87 & 85 & 94 & 85 & 43 & 73 \\
\raisebox{-0.1\height}{\includegraphics[height=1em]{imgs/openai-logo.png}} GPT-5 mini         & 68 & 68 & 20 & 84 & 91 & 84 & 84 & 39 & 75 \\
\raisebox{-0.1\height}{\includegraphics[height=1em]{imgs/openai-logo.png}} GPT-5 nano         & 59 & 42 & 8 & 78 & 84 & 79 & 79 & 37 & 68 \\
\hdashline
\raisebox{-0.1\height}{\includegraphics[height=1em]{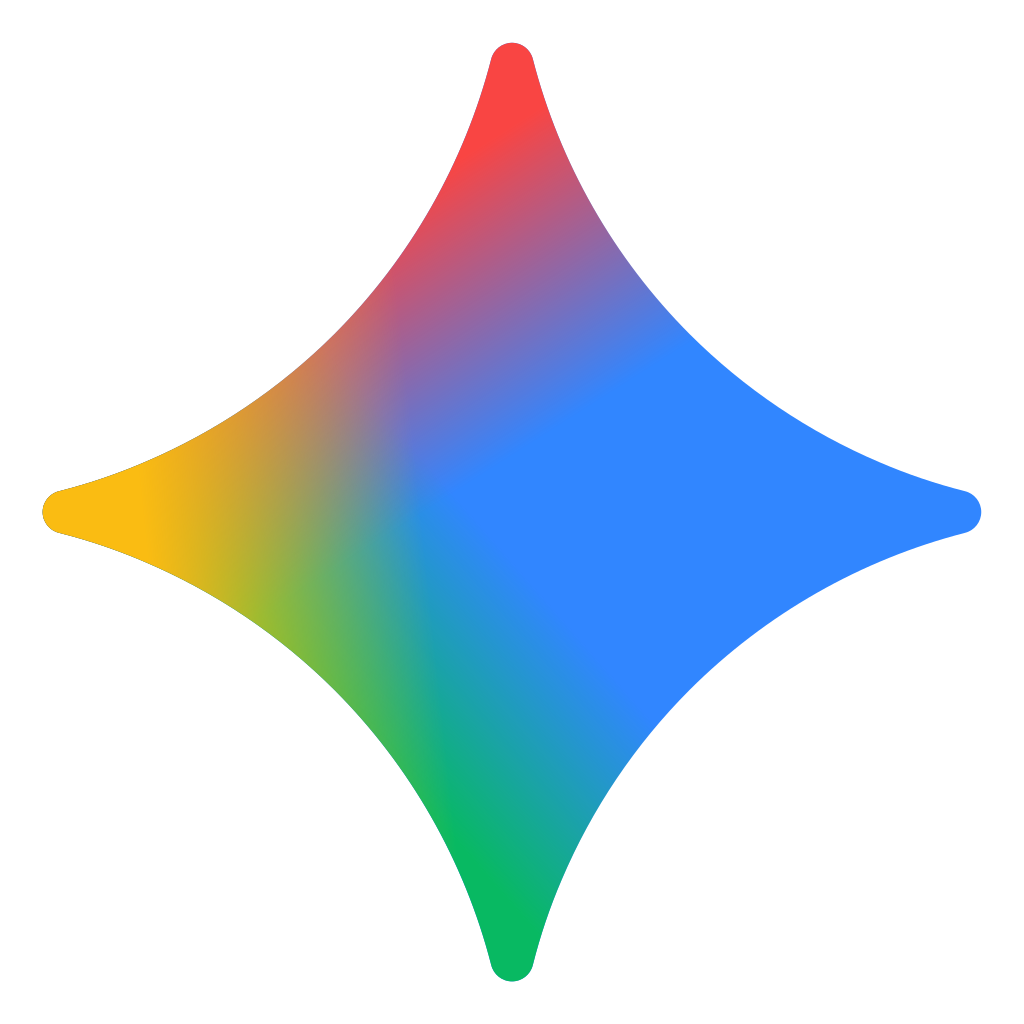}} Gemini-3.0 Pro     & 75 & 71 & 37 & 90 & 91 & 96 & 92 & 56 & 70 \\
\raisebox{-0.1\height}{\includegraphics[height=1em]{imgs/gemini-logo.png}} Gemini-3.0 Flash   & 75 & 66 & 35 & 89 & 90 & 97 & 91 & 51 & 78 \\
\raisebox{-0.1\height}{\includegraphics[height=1em]{imgs/gemini-logo.png}} Gemini-2.5 Pro     & 65 & 66 & 21 & 86 & 84 & 88 & 80 & 43 & 49 \\
\raisebox{-0.1\height}{\includegraphics[height=1em]{imgs/gemini-logo.png}} Gemini-2.5 Flash   & 58 & 57 & 13 & 84 & 79 & 78 & 63 & 41 & 52 \\
\raisebox{-0.1\height}{\includegraphics[height=1em]{imgs/gemini-logo.png}} Gemini-2.0 Flash   & 47 & 45$^*$ & 8$^*$ & 77 & 68$^*$ & 73 & 39$^*$ & 35$^*$ & 30$^*$ \\
\hdashline
\raisebox{-0.1\height}{\includegraphics[height=1em]{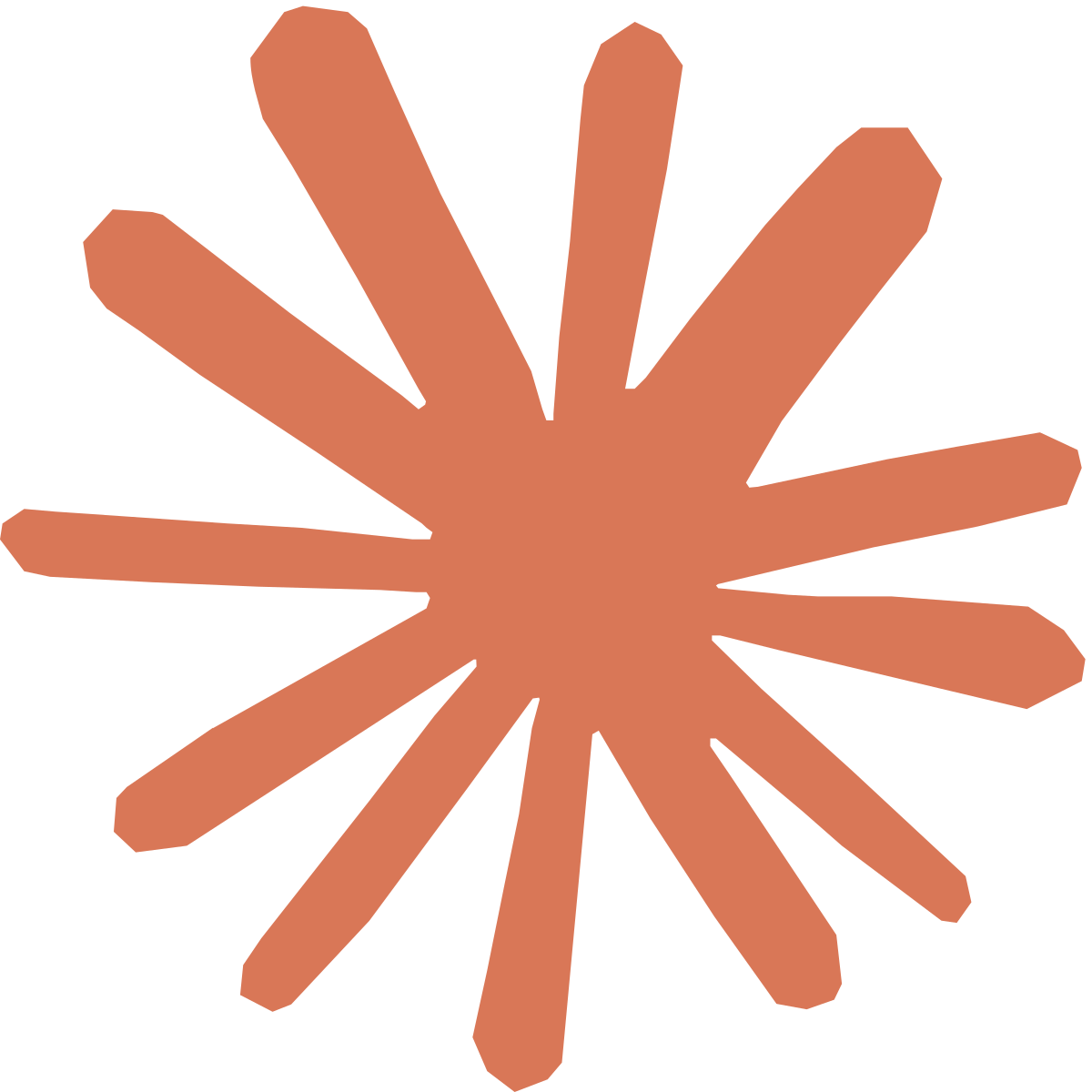}} Claude Sonnet 4.5  & 55 & 66 & 7 & 88 & 83 & 37 & 71 & 43 & 43 \\
\raisebox{-0.1\height}{\includegraphics[height=1em]{imgs/claude-logo.png}} Claude Sonnet 4  & 47 & 62$^*$ & 5$^*$ & 87 & 75 & 21 & 56$^*$ & 38$^*$ & 35$^*$ \\
\raisebox{-0.1\height}{\includegraphics[height=1em]{imgs/claude-logo.png}} Claude 3.7 Sonnet & 42 & 58$^*$ & 2$^*$ & 81 & 67 & 12 & 57 & 32$^*$ & 30$^*$ \\
\bottomrule
\end{tabular}
}
\vspace{1mm}
\raggedright
\footnotesize{$^*$ Estimated or averaged from reported range.}
\end{table*}

Our reconstructed Intelligence Index values should be interpreted with appropriate caution.  
First, several evaluations, particularly long-context and agentic workflow tasks, contain nondeterministic components that may vary slightly across implementations.
Second, for models without public API support for large-context evaluation (e.g., ``non-reasoning'' checkpoints), our long-context estimates represent upper-bound approximations based on available context windows and internal model behavior.  
Third, Artificial Analysis maintains private test variants and additional filtering procedures that cannot be fully reproduced. Thus, our estimates provide a methodologically aligned but not officially verified extension.

\begin{table}[t]
\caption{Aggregate out-of-sample validation metrics across three held-out models (GPT-5.2, Gemini-3.0 Pro, Gemini-3.0 Flash) on BrowseComp-Plus, all with Intelligence Index = 75. The scaling equation achieves well-calibrated MAS predictions while systematically over-predicting single-agent performance.}
\label{tab:validation_summary}
\centering
\begin{tabular}{l c c}
\toprule
\textbf{Metric} & \textbf{Value} & \textbf{Note} \\
\midrule
Overall MAE & 0.077 & 15 predictions \\
MAS-only MAE & 0.061 & 12 predictions \\
SAS-only MAE & 0.138 & 3 predictions \\
\midrule
Overall MAPE & 19.9\% & --- \\
MAS-only MAPE & 15.3\% & --- \\
\midrule
Findings Validated & 11/15 & 73\% \\
Findings Partial & 2/15 & 13\% \\
\bottomrule
\end{tabular}
\end{table}

\begin{table*}[t]
\caption{Architecture-wise prediction accuracy for three held-out models on BrowseComp-Plus. All models share Intelligence Index = 75. MAS predictions are well-calibrated (average error 6.1\%), while SAS shows systematic over-prediction due to linear extrapolation beyond the training range.}
\label{tab:architecture_predictions}
\footnotesize
\centering
\begin{tabular}{l ccc ccc ccc}
\toprule
& \multicolumn{3}{c}{\textbf{GPT-5.2}} & \multicolumn{3}{c}{\textbf{Gemini-3.0 Pro}} & \multicolumn{3}{c}{\textbf{Gemini-3.0 Flash}} \\
\cmidrule(lr){2-4} \cmidrule(lr){5-7} \cmidrule(lr){8-10}
\textbf{Architecture} & Pred. & Actual & Error & Pred. & Actual & Error & Pred. & Actual & Error \\
\midrule
SAS & 0.521 & 0.450 & $+$15.8\% & 0.521 & 0.360 & $+$44.7\% & 0.521 & 0.340 & $+$53.2\% \\
MAS-Centralized & 0.480 & 0.480 & $+$0.0\% & 0.480 & 0.440 & $+$9.1\% & 0.480 & 0.480 & $+$0.0\% \\
MAS-Decentralized & 0.496 & 0.480 & $+$3.3\% & 0.496 & 0.500 & $-$0.8\% & 0.496 & 0.400 & $+$24.0\% \\
MAS-Independent & 0.413 & 0.350 & $+$18.0\% & 0.413 & 0.400 & $+$3.2\% & 0.413 & 0.360 & $+$14.7\% \\
MAS-Hybrid & 0.560 & 0.390 & $+$43.6\% & 0.560 & 0.440 & $+$27.3\% & 0.560 & 0.400 & $+$40.0\% \\
\midrule
\textbf{MAE (Overall)} & \multicolumn{3}{c}{0.064} & \multicolumn{3}{c}{0.068} & \multicolumn{3}{c}{0.098} \\
\textbf{MAE (MAS only)} & \multicolumn{3}{c}{0.062} & \multicolumn{3}{c}{0.044} & \multicolumn{3}{c}{0.077} \\
\bottomrule
\end{tabular}
\end{table*}

\begin{table*}[t]
\caption{Validation of key findings from Section~\ref{sec:results} across three held-out models. Three findings generalize universally (\textcolor{green!60!black}{\ding{51}}), while two show model-family-specific patterns. Per-model validation rates are 4/5, 4/5, and 3/5 respectively.}
\label{tab:findings_validation}
\footnotesize
\centering
\begin{tabularx}{\textwidth}{X c c c}
\toprule
\textbf{Finding} & \textbf{GPT-5.2} & \textbf{Gemini-3.0 Pro} & \textbf{Gemini-3.0 Flash} \\
\midrule
Capability Ceiling (higher $P_{SA}$ correlates with diminishing MAS returns) 
& \textcolor{green!60!black}{\ding{51}} & \textcolor{green!60!black}{\ding{51}} & \textcolor{green!60!black}{\ding{51}} \\

Independent MAS Degradation (Independent underperforms SAS)
& \textcolor{green!60!black}{\ding{51}} & \textcolor{red!60!black}{\ding{55}}$^\dagger$ & Partial$^\dagger$ \\

Optimal Architecture (Centralized/Decentralized excel)
& \textcolor{green!60!black}{\ding{51}} & \textcolor{green!60!black}{\ding{51}} & \textcolor{green!60!black}{\ding{51}} \\

Hybrid Overhead (515\% overhead limits performance)
& \textcolor{green!60!black}{\ding{51}} & \textcolor{green!60!black}{\ding{51}} & \textcolor{green!60!black}{\ding{51}} \\

BrowseComp-Plus Pattern (Decentralized $\geq$ Centralized)
& Partial$^\ddagger$ & \textcolor{green!60!black}{\ding{51}} & \textcolor{red!60!black}{\ding{55}}$^\S$ \\
\midrule
\textbf{Total Validated} & \textbf{4/5} & \textbf{4/5} & \textbf{3/5} \\
\bottomrule
\end{tabularx}
\vspace{-4pt}
\begin{flushleft}
\footnotesize{$^\dagger$ Gemini models show Independent MAS matching or outperforming SAS (+11.1\% for Pro, +5.9\% for Flash), contrary to GPT-5.2 ($-$22.2\%). This may reflect model-family-specific agentic capabilities.} \\
\footnotesize{$^\ddagger$ GPT-5.2 shows convergence (both 0.48); main results predicted Decentralized advantage.} \\
\footnotesize{$^\S$ Gemini-3.0 Flash shows reversed pattern: Centralized (0.48) $>$ Decentralized (0.40).}
\end{flushleft}
\end{table*}

\begin{table}[t]
\caption{Model family differences despite identical Intelligence Index (75). Single-agent performance varies substantially across vendors, while best MAS performance converges, suggesting multi-agent architectures may compensate for single-agent limitations.}
\label{tab:model_family_differences}
\centering
\begin{tabular}{l c c c}
\toprule
\textbf{Model} & $P_{\text{SA}}$ & Best MAS & MAS Gain \\
\midrule
GPT-5.2 & 0.45 & 0.48 & $+$6.7\% \\
Gemini-3.0 Pro & 0.36 & 0.50 & $+$38.9\% \\
Gemini-3.0 Flash & 0.34 & 0.48 & $+$41.2\% \\
\bottomrule
\end{tabular}
\vspace{2pt}
\begin{flushleft}
\footnotesize{Note: All models share Intelligence Index = 75, yet exhibit different single-agent performance. This suggests Intelligence Index may not be directly comparable across model families.}
\end{flushleft}
\end{table}

\section{Out-of-Sample Validation}
\label{sec:validation}

To assess the generalizability of our scaling equation beyond the training distribution, we evaluate on three held-out models: GPT-5.2, Gemini-3.0 Pro, and Gemini-3.0 Flash. All three models share Intelligence Index = 75, representing extrapolation beyond our training range (Index 42--71) by approximately 5.6\%. Table~\ref{tab:validation_summary} summarizes aggregate validation metrics. Across 15 architecture-model combinations, the scaling equation achieves MAE = 0.077. MAS predictions are better calibrated (MAE = 0.061) than SAS predictions (MAE = 0.138), consistent with systematic over-prediction of single-agent performance when extrapolating beyond the training range.

\paragraph{Qualitative Findings Validation.}
Table~\ref{tab:findings_validation} evaluates whether the five key findings from Section~\ref{sec:results} generalize to held-out models. Across three models, 11 of 15 finding-model pairs validate (73\%), with two additional partial validations. Three findings generalize universally: (1) the capability ceiling effect persists across all models, (2) Centralized or Decentralized architectures achieve optimal performance, and (3) Hybrid overhead limits relative performance. Two findings show model-family-specific behavior: Independent MAS degradation validates only for GPT-5.2 but not for Gemini models, and the BrowseComp pattern (Decentralized $>$ Centralized) varies across model families.

\paragraph{Model Family Differences.}
An informative comparison arises from models with identical Intelligence Index (Table~\ref{tab:model_family_differences}). Despite Index = 75, single-agent performance varies substantially: GPT-5.2 achieves $P_{\text{SA}} = 0.45$, while Gemini-3.0 Pro and Gemini-3.0 Flash achieve 0.36 and 0.34, respectively. However, best MAS performance converges ($0.48$--$0.50$), suggesting that multi-agent architectures may compensate for single-agent limitations. Consequently, MAS gains are substantially higher for Gemini models (+38.9\% and +41.2\%) compared to GPT-5.2 (+6.7\%). This implies that Intelligence Index, while predictive within model families, may not be directly comparable across vendors, a limitation for cross-family extrapolation that future scaling laws should address.

\textbf{Architecture Selection Accuracy.} The scaling equation predicts Hybrid as optimal for all three models
($\hat{P}_{\text{Hybrid}} = 0.560$), yet empirically Centralized and Decentralized architectures achieve superior performance. This
discrepancy reflects two factors: (1) linear extrapolation of the intelligence effect ($\hat{\beta}_I = 0.126$) beyond its training
range, and (2) the model's failure to capture Hybrid's disproportionate overhead penalty at high capability levels. 
The equation systematically over-predicts Hybrid performance (mean error $+37.0\%$) while achieving reasonable calibration for Centralized (mean error $+3.0\%$) and Decentralized (signed mean error $+8.8\%$). These results suggest that while the scaling equation provides well-calibrated predictions for moderate-overhead architectures, high-overhead configurations like Hybrid require architecture-specific corrections when extrapolating to frontier models.

\section{Domain Complexity}
\label{appendix:domain_complexity}

\textcolor{black}{We characterize domain complexity through an ordinal score $D \in [0,1]$ that captures the degree of sequential interdependence and empirical difficulty across evaluated benchmarks. This characterization enables systematic analysis of when multi-agent coordination yields performance benefits versus incurring prohibitive overhead.}

\subsection{Complexity Score Assignment}

\textcolor{black}{Domain complexity $D \in [0, 1]$ is assigned based on three empirical task properties, each normalized to $[0,1]$:}

\begin{itemize}

    \item \textcolor{black}{\textbf{Sequential Interdependence.} The degree to which task completion requires strictly ordered reasoning steps. Tasks with parallelizable subtask structure (e.g., Finance Agent) score low, while tasks requiring sequential constraint satisfaction (e.g., PlanCraft) score high.}

    \item \textcolor{black}{\textbf{State-Space Complexity.} The extent of dynamic state evolution during task execution. Tasks with static or slowly evolving states score low, while tasks requiring tracking of rapidly changing environments (e.g., BrowseComp-Plus) score high.}

    \item \textcolor{black}{\textbf{Coordination Overhead Sensitivity.} Empirically observed degradation under multi-agent coordination relative to single-agent baselines, reflecting how much the task's structure penalizes inter-agent communication overhead.}
\end{itemize}

\textcolor{black}{The final score reflects the overall empirical difficulty profile, calibrated against observed MAS performance patterns across all configurations.}

\subsection{Domain Characterisation}

Table~\ref{tab:domain_complexity} summarises the complexity scores and defining characteristics of each benchmark.

\begin{table*}[h]
\centering
\caption{Domain complexity scores and task characteristics.}
\label{tab:domain_complexity}
\begin{tabular}{lcp{11.3cm}}
\toprule
\textbf{Domain} & \textbf{$D$} & \textbf{Characteristics} \\
\midrule
Workbench & 0.000 & Minimal sequential constraints; well-structured procedural reasoning with clear subtask boundaries; low coordination requirements \\
Finance Agent & 0.407 & Moderate decomposability; structured domains amenable to localised agent reasoning \\
PlanCraft & 0.419 & High sequential dependencies; constraint satisfaction requiring ordered reasoning steps \\
BrowseComp-Plus & 0.839 & Dynamic state evolution; complex visuospatial reasoning with interaction-heavy environments \\
\textcolor{black}{SWE-bench Verified} & \textcolor{black}{0.255} & \textcolor{black}{Decomposable software engineering tasks; multi-step codebase exploration with test feedback; high tool count (7)} \\
\textcolor{black}{Terminal-Bench} & \textcolor{black}{0.414} & \textcolor{black}{Diverse CLI tasks with varying difficulty; Docker-based environments; low tool count (2)} \\
\bottomrule
\end{tabular}
\end{table*}

\subsection{Critical Threshold}

Our analysis identifies a critical complexity threshold at $D \approx 0.40$. Below this threshold, multi-agent architectures yield net positive returns through effective task decomposition and parallel reasoning. Above this threshold, coordination overhead consumes computational resources otherwise allocated to reasoning, resulting in performance degradation. This finding suggests that the suitability of multi-agent approaches is fundamentally constrained by domain-intrinsic properties rather than architectural sophistication alone.

\section{Datasets}

\textcolor{black}{We evaluate our agent systems across six agentic benchmarks requiring multi-step reasoning and tool interaction. Each dataset emphasizes different aspects of agentic behavior: information retrieval, domain expertise, planning, task decomposition, software engineering, and terminal-based task execution.}

\paragraph{Finance Agent.}

We use the Finance Agent benchmark \citep{bigeard2025finance}, comprising 50 finance questions requiring domain expertise and multi-step analysis. Tasks include earnings analysis, financial metric calculations, and market trend interpretation. Each instance includes expert-provided rubrics for structured evaluation. Questions typically require 15-30 minutes of expert time, indicating substantial complexity. 

\paragraph{BrowseComp Plus.}
BrowseComp Plus \citep{chen2025browsecomp} contains 100 web browsing tasks requiring multi-website information synthesis. Tasks include comparative analysis, fact verification, and
multi-source research across the web. Each instance requires agents to navigate multiple websites, extract relevant details, and synthesize findings. The dataset uses LLM-based evaluation comparing agent responses against ground truth answers with confidence scoring.

\paragraph{WorkBench.}
WorkBench \citep{styles2024workbench} evaluates business task automation through function calling sequences. The dataset covers five domains: analytics, calendar management, email
operations, project management, and customer relationship management. Success requires executing correct tool sequences to accomplish realistic business workflows. Evaluation follows outcome-centric assessment, measuring exact match between predicted and expected function call sequences. The dataset supports 100 distinct business scenarios with tolerance for minor date variations.

\paragraph{Plancraft.}
Plancraft \citep{dagan2024plancraft} focuses on sequential planning in Minecraft environments. Agents must craft target items by determining optimal action sequences using available inventory and crafting recipes. Tasks require multi-step reasoning about dependencies, resource management, and action ordering. The dataset uses environment-determined success
metrics based on successful item crafting within step limits. We use the plancraft-test subset containing focused planning challenges.

\textcolor{black}{\paragraph{SWE-bench Verified.}
SWE-bench Verified \citep{jimenez2023swe} evaluates software engineering capabilities through real-world GitHub issue resolution. Each instance provides a repository snapshot and issue description; agents must produce a patch that resolves the issue and passes the repository's test suite. The benchmark provides 7 tools including bash execution, file editing, directory navigation, search, and test execution, requiring multi-step codebase exploration, hypothesis generation, and iterative debugging. We evaluate on a 20-instance subset selected via deterministic shuffle (seed 42) from the 500-instance verified split, balancing computational cost with coverage across repository diversity.}

\textcolor{black}{\paragraph{Terminal-Bench.}
Terminal-Bench \citep{merrill2026terminal} evaluates CLI task execution across diverse system administration, security, machine learning, and debugging scenarios. Each instance specifies a terminal task with a Docker-based evaluation environment and objective success criteria. Agents interact through 2 tools (bash command execution and answer submission), requiring sustained environmental interaction under varying time limits. We evaluate on the first 20 instances from the 86-instance benchmark, covering tasks ranging from file manipulation and network configuration to model training and system diagnostics.}

\section{Implementation Details}

\subsection{Technical Infrastructure}

\textcolor{black}{Our implementation uses LiteLLM (\url{https://www.litellm.ai/}) for unified API access across model providers and LangChain (\url{https://www.langchain.com/}) for agent orchestration and tool integration.} LiteLLM provides standardized interfaces for OpenAI, Gemini, and Anthropic models, enabling seamless model switching and comparison. LangChain facilitates tool binding, conversation management, and structured prompting.

\paragraph{API Integration.}

We access LLMs through provider-specific APIs: OpenAI API for GPT models (gpt-5, gpt-5-mini, gpt-5-nano), GenAI API for Gemini models (gemini-2.5-pro, gemini-2.5-flash, gemini-2.0-flash), and Anthropic API for Claude models (claude-4.5-sonnet, claude-4.0-sonnet, claude-3.7-sonnet). Our implementation includes intelligent API key rotation across multiple keys per provider to handle rate limiting and quota management. Context window management automatically truncates conversation history when token limits are approached.

\paragraph{Tool Environment.}
Each dataset defines its tool ecosystem through environment configurations. \textcolor{black}{Tools include web search (Tavily, \url{https://tavily.com/}), code execution} (Python REPL), mathematical operations, and task
completion markers. Tool definitions use LangChain's BaseTool interface with structured input schemas and execution methods. Tools are dynamically bound to LLM instances using
function calling capabilities when available.

\subsection{Agent Configuration}

\paragraph{Architecture Parameters.}
Single agents use maximum 10 iterations per instance. Independent multi-agent systems deploy 3 agents with synthesis-only coordination. Centralized systems employ 3 sub-agents with 1 orchestrator across maximum 5 orchestration rounds, with 3 iterations per agent per round. Decentralized systems run 3 agents through 3 debate rounds with 3 iterations per round. Hybrid systems combine centralized orchestration with limited peer communication phases.

\paragraph{Heterogeneous Models.}
Our framework supports heterogeneous configurations where different agent roles use different models. Orchestrators can use high-capability models (e.g., GPT-5) while sub-agents use
efficient models (e.g., Gemini-2.0 Flash). The LLMConfig class manages model assignment with automatic LLM instance creation for each agent role. Decentralized systems can assign
different models to different workers for diversity.

\subsection{Prompt Compilation System}

We implement a structured prompting system supporting named templates and variable interpolation. Prompts are defined in YAML files with base templates and role-specific extensions.
The compilation process performs template variable replacement using double-brace syntax ({{variable}}) and supports conditional template selection based on agent type and conversation state.

\paragraph{Dataset Integration.}
Each dataset provides shared prompt templates containing task-specific instructions and examples. Dataset instances contribute prompt variables including problem descriptions,
context, and constraints. The prompt compilation system merges agent prompts with dataset templates, ensuring consistent instruction delivery across architectures while maintaining task specificity.

\subsection{Evaluation Methodology}

\paragraph{Sample Sizes.}
\textcolor{black}{We evaluate on dataset subsets balancing computational cost with statistical significance: Finance Agent (50 instances), BrowseComp Plus (100 instances), WorkBench (100 instances), Plancraft (100 instances), SWE-bench Verified (20 instances, deterministic shuffle with seed 42 from 500), and Terminal-Bench (20 instances, first-$N$ from 86). The smaller subsets for SWE-bench Verified and Terminal-Bench reflect the computational cost of Docker-based evaluation environments; bootstrap 95\% confidence intervals are reported in Table \ref{tab:s6_bootstrap}.} Instance selection ensures representative coverage of task types and difficulty levels within each benchmark.

\paragraph{Restrictions and Controls.}
All experiments use identical tool interfaces and observation structures across architectures to eliminate external feedback confounds. Context window management applies consistent truncation policies. API rate limiting and retry mechanisms ensure fair resource allocation. Evaluation uses frozen model weights without fine-tuning to measure architectural effects independently of model optimization.

\subsection{Information Gain Computation}
\label{appendix:info_gain}

Information gain $\Delta\mathcal{I}$ quantifies the reduction in task uncertainty achieved through agent coordination. We estimate this via Bayesian posterior variance reduction:
\begin{equation}
\Delta\mathcal{I} = \frac{1}{2}\log\frac{\text{Var}[Y | \mathbf{s}_{\text{pre}}]}{\text{Var}[Y | \mathbf{s}_{\text{post}}]},
\label{eq:info_gain}
\end{equation}
where $Y \in \{0,1\}$ is the task success indicator, $\mathbf{s}_{\text{pre}}$ is the agent's state representation before coordination (initial reasoning trace), and $\mathbf{s}_{\text{post}}$ is the state after coordination (final aggregated output). Variances are estimated via Monte Carlo sampling: we generate $K=10$ reasoning traces per state using temperature $\tau=0.7$ and compute empirical variance of predicted success probabilities. For binary outcomes, this reduces to:
\begin{equation}
\text{Var}[Y | \mathbf{s}] = \hat{p}(\mathbf{s})(1 - \hat{p}(\mathbf{s})),
\end{equation}
where $\hat{p}(\mathbf{s})$ is the mean predicted success probability across samples.

\startcontents[appendices]

\begin{figure}[ht!]
      \centering
      \includegraphics[width=0.9\linewidth]{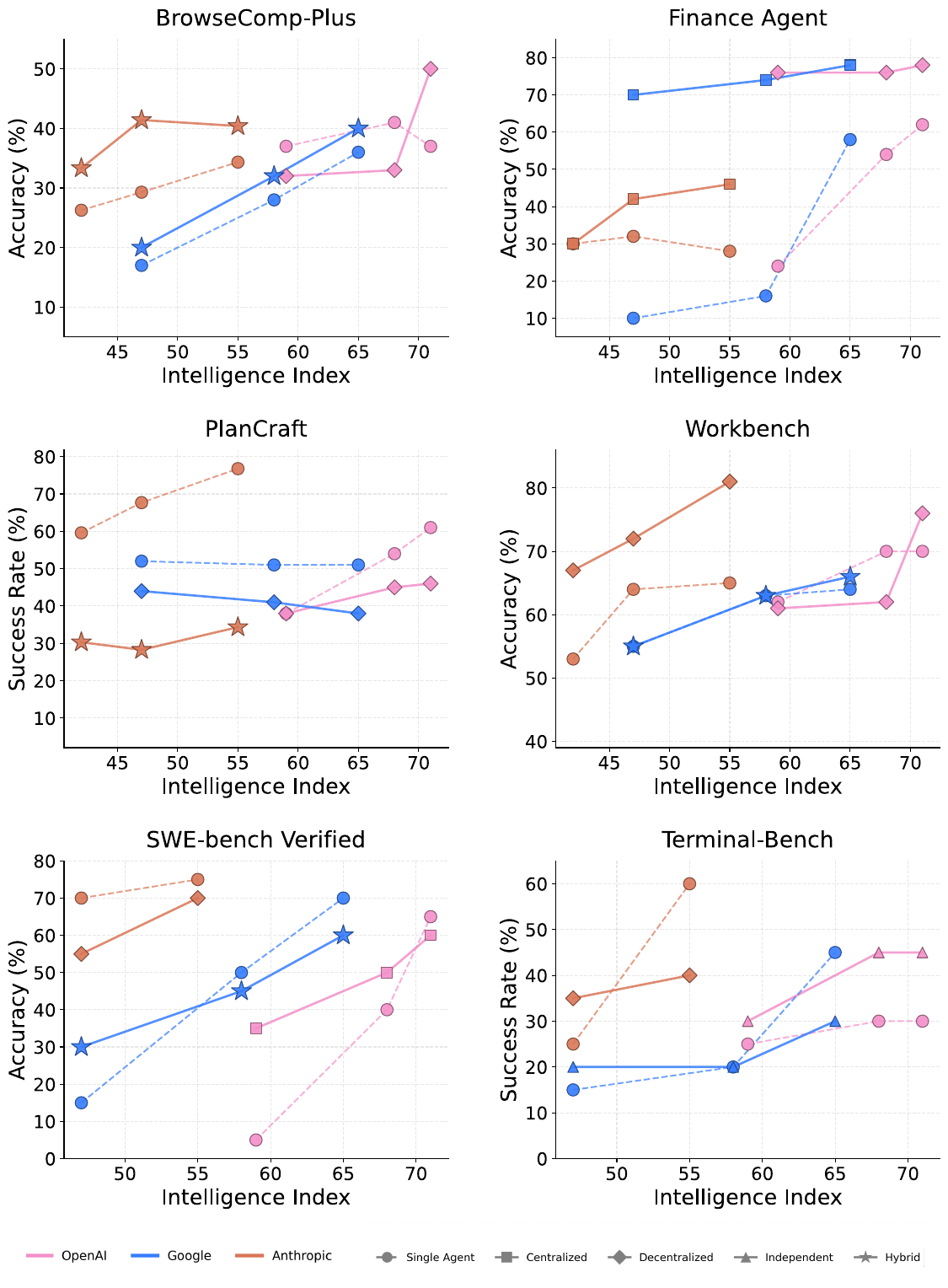}
      \caption{\textcolor{black}{\textbf{Agent scaling dynamics across model capability.}
  Performance across six benchmarks show best-performing multi-agent system versus single-agent baselines by Intelligence Index. OpenAI and Google show cooperative scaling in structured tasks (Finance Agent, Workbench). Anthropic models show diminished or negative returns in open-ended environments (PlanCraft, BrowseComp-Plus). SWE-bench Verified and Terminal-Bench show patterns consistent with the capability-saturation threshold: SWE-bench (high SAS baselines) shows limited MAS gains, while Terminal-Bench (lower baselines) shows mixed results with low tool count limiting coordination benefits.}}
      \label{fig:main_exp}
  \end{figure}

% \textcolor{black}{
\section{SWE-bench Verified and Terminal-Bench Results}

\begin{table}[h]
\centering
\small
\caption{Heterogeneous vs. homogeneous agent configurations on BrowseComp-Plus. Centralized heterogeneous configurations uniformly underperform the stronger model's homogeneous baseline; decentralized configurations show modest gains largely attributable to the stronger constituent model.}
\begin{tabular}{llcc}
\toprule
\textbf{Architecture} & \textbf{Configuration} & \textbf{Accuracy} & \textbf{$\Delta$ vs. Homo (strong)} \\
\midrule
Centralized & GPT-5 orch + GPT-5-nano sub & 0.19 & $-$0.15 \\
Centralized & GPT-5-nano orch + GPT-5 sub & 0.21 & $-$0.13 \\
Centralized & Sonnet-4.5 orch + Sonnet-3.7 sub & 0.37 & $-$0.06 \\
Centralized & Sonnet-3.7 orch + Sonnet-4.5 sub & 0.42 & $-$0.01 \\
Centralized & Gemini-2.5-Pro orch + 2.0-Flash sub & 0.18 & $-$0.19 \\
Centralized & Gemini-2.0-Flash orch + 2.5-Pro sub & 0.23 & $-$0.14 \\
Centralized & Gemini-2.5-Pro orch + GPT-5 sub & 0.17 & $-$0.20 \\
\midrule
Decentralized & 2 GPT-5 + 1 GPT-5-nano & 0.56 & $+$0.06 \\
Decentralized & 1 GPT-5 + 2 GPT-5-nano & 0.51 & $+$0.01 \\
Decentralized & 2 Sonnet-4.5 + 1 Sonnet-3.7 & 0.45 & $+$0.02 \\
Decentralized & 1 Sonnet-4.5 + 2 Sonnet-3.7 & 0.48 & $+$0.05 \\
Decentralized & 2 Gemini-2.5-Pro + 1 2.0-Flash & 0.47 & $+$0.04 \\
Decentralized & 1 Gemini-2.5-Pro + 2 2.0-Flash & 0.37 & $-$0.06 \\
\bottomrule
\end{tabular}
\label{tab:s2_heterogeneous}
\end{table}

\begin{table}[h]
\centering
\small
\caption{Regression fit under alternative capability metrics (6-benchmark model, $N=260$). The Agentic Capability Index (ACI) achieves the best cross-validated fit with zero finding reversals.}
\begin{tabular}{lccc}
\toprule
\textbf{Capability Metric} & $R^2_{\text{train}}$ & $R^2_{\text{CV}}$ & \textbf{AIC} \\
\midrule
Intelligence Index & 0.463 & 0.373 & $-$236.3 \\
Agentic Capability Index (ACI) & \textbf{0.481} & \textbf{0.413} & $\mathbf{-244.8}$ \\
Per-dataset SA Baseline & 0.372 & 0.247 & $-$202.0 \\
\bottomrule
\end{tabular}
\label{tab:s3_aci}
\end{table}

\begin{table}[h]
\centering
\small
\caption{Naive vs. cluster-robust $p$-values for key predictors (6-benchmark model, $N=260$, $G=6$ clusters). Predictors varying at the dataset level show substantial SE inflation under cluster-robust inference.}
\begin{tabular}{lccl}
\toprule
\textbf{Predictor} & \textbf{Naive $p$} & \textbf{Robust $p$} & \textbf{Status} \\
\midrule
\texttt{single\_agent\_baseline} & 0.001 & \textbf{0.004} & Survives \\
\texttt{error\_x\_baseline} & 0.022 & \textbf{0.030} & Survives \\
\texttt{log\_tools} & $<$0.001 & 0.172 & Inflated \\
\texttt{intelligence\_centered} & 0.008 & 0.059 & Inflated \\
\texttt{efficiency\_x\_tools} & 0.002 & 0.205 & Inflated \\
\texttt{baseline\_x\_agents} & 0.004 & 0.105 & Inflated \\
\bottomrule
\end{tabular}
\label{tab:s4_cluster}
\end{table}

\begin{table}[h]
\centering
\small
\caption{Holm--Bonferroni multiple comparison correction (19 hypotheses, $\alpha=0.05$). Three predictors survive family-wise error rate correction.}
\begin{tabular}{lccc}
\toprule
\textbf{Predictor} & \textbf{Raw $p$} & \textbf{Holm $p$} & \textbf{Bonferroni $p$} \\
\midrule
\texttt{log\_tools} & $<$0.001 & $<$0.001 & $<$0.001 \\
\texttt{single\_agent\_baseline} & 0.001 & 0.018 & 0.019 \\
\texttt{efficiency\_x\_tools} & 0.002 & 0.026 & 0.030 \\
\midrule
\texttt{intelligence\_centered} & 0.008 & 0.056 & 0.066 \\
\texttt{baseline\_x\_agents} & 0.004 & 0.084 & 0.106 \\
\bottomrule
\end{tabular}
\label{tab:s5_holm}
\end{table}

\begin{table}[h]
\centering
\small
\caption{SWE-bench Verified and Terminal-Bench resolution rates with 95\% bootstrap confidence intervals ($n=20$ instances, 10,000 resamples). Values shown as point estimate [\,lower, upper\,].}
\begin{tabular}{l|ccccc}
\toprule
& \textbf{Single} & \textbf{Centralized} & \textbf{Decentralized} & \textbf{Hybrid} & \textbf{Independent} \\
\midrule
\multicolumn{6}{c}{\textit{SWE-bench Verified}} \\
\midrule
gemini-2.0-flash & 15 [0, 30] & 25 [10, 45] & 25 [5, 45] & 30 [10, 50] & 40 [20, 60] \\
gemini-2.5-flash & 50 [30, 70] & 35 [15, 55] & 35 [15, 55] & 45 [25, 65] & 40 [20, 60] \\
gemini-2.5-pro & 70 [50, 90] & 55 [35, 75] & 45 [25, 65] & 60 [40, 80] & 50 [30, 70] \\
gemini-3-flash & 80 [60, 95] & 75 [55, 95] & 80 [60, 95] & 75 [55, 95] & 60 [40, 80] \\
gpt-5-nano & 5 [0, 15] & 35 [15, 55] & 35 [15, 55] & 30 [10, 50] & 25 [10, 45] \\
gpt-5-mini & 40 [20, 60] & 50 [30, 70] & 30 [10, 50] & 50 [30, 70] & 45 [25, 65] \\
gpt-5 & 65 [45, 85] & 60 [40, 80] & 70 [50, 90] & 55 [35, 75] & 55 [35, 75] \\
claude-sonnet-4 & 70 [50, 90] & 50 [30, 70] & 55 [35, 75] & 45 [25, 65] & 30 [10, 50] \\
claude-sonnet-4-5 & 75 [55, 95] & 70 [50, 90] & 70 [50, 90] & 70 [50, 90] & 55 [35, 75] \\
\midrule
\multicolumn{6}{c}{\textit{Terminal-Bench}} \\
\midrule
gemini-2.0-flash & 15 [0, 30] & 15 [0, 30] & 5 [0, 15] & 5 [0, 15] & 20 [5, 40] \\
gemini-2.5-flash & 20 [5, 40] & 25 [10, 45] & 25 [10, 45] & 25 [10, 45] & 20 [5, 40] \\
gemini-2.5-pro & 45 [25, 65] & 10 [0, 25] & 30 [10, 50] & 25 [10, 45] & 30 [10, 50] \\
gemini-3-flash & 60 [40, 80] & 50 [30, 70] & 55 [35, 75] & 45 [25, 65] & 50 [30, 70] \\
gpt-5-nano & 25 [10, 45] & 20 [5, 40] & 25 [10, 45] & 20 [5, 40] & 30 [10, 50] \\
gpt-5-mini & 30 [10, 50] & 15 [0, 30] & 30 [10, 50] & 20 [5, 40] & 45 [25, 65] \\
gpt-5 & 30 [10, 50] & 45 [25, 65] & 45 [25, 65] & 50 [30, 70] & 45 [25, 65] \\
claude-sonnet-4 & 25 [10, 45] & 25 [5, 45] & 35 [15, 55] & 25 [10, 45] & 35 [15, 55] \\
claude-sonnet-4-5 & 60 [40, 80] & 45 [25, 65] & 40 [20, 60] & 50 [30, 70] & 40 [20, 60] \\
\bottomrule
\end{tabular}
\label{tab:s6_bootstrap}
\end{table}
% }
\end{document}